\documentclass{article}

\usepackage[final]{corl_2026} 

\usepackage{cite}
\usepackage{amsmath,amssymb,amsfonts}
\usepackage{algorithmic}
\usepackage{graphicx}
\usepackage{textcomp}
\usepackage{url}
\usepackage{footnote}
\usepackage{graphics}
\usepackage{subfigure}

\usepackage{multirow}
\usepackage{booktabs}
\usepackage{tabularx}
\usepackage{hyperref}
\usepackage{gensymb}
\usepackage{wrapfig}
\usepackage{enumitem}

\usepackage{xurl}

\title{Cyclops: LiDAR as a Camera That Dreams in Color}

\author{
    \textbf{Wei Gao$^{1}$,
    Jian Shu$^{2}$,
    Mingle Zhao$^{1}$,
    Maani Ghaffari$^{3}$,
    David Kong$^{4}$,} \\
    \textbf{Chengzhong Xu$^{1}$,
    and Hui Kong$^{1,*}$} \\ [5pt]
    $^{1}$University of Macau, $^{2}$The Hong Kong University of Science and Technology (Guangzhou), \\
    $^{3}$University of Michigan, $^{4}$Independent Researcher \\ [5pt]
    $^{*}$Corresponding Author \\ [5pt]
    \href{https://github.com/IMRL/Cyclops}{https://github.com/IMRL/Cyclops}
}

\begin{document}
\maketitle


\begin{abstract}
    Conventionally, robotic perception relies heavily on cameras due to the rich semantic texture they provide. However, their performance degrades significantly in low-light or high-dynamic-range environments. Conversely, while Light Detection and Ranging (LiDAR) captures illumination-invariant geometric and intensity properties, the resulting data are typically single-channel and sparse, creating a significant modality gap when applying vision models pre-trained on RGB datasets. In this paper, we propose \textit{Cyclops}, a framework that translates sparse Non-Repetitive Scanning LiDAR (NRS-LiDAR) intensity into RGB video, enabling camera-free inference for all-day and all-illumination perception tasks. Our approach first converts sparse LiDAR intensity projections into dense representations via a frozen pre-trained densification module, serving as a geometrically rich source condition. The dense intensity latent is then transported toward the target RGB distribution through Latent Bridge Matching (LBM) with a learned velocity field in a few ODE integration steps. To mitigate inter-frame flickering, we inject prior-frame context via temporal attention layers and further formulate the velocity field as a policy optimized by a differentiable terminal reward that encourages terminal fidelity through backpropagation along the ODE trajectory. Extensive experiments demonstrate that the predicted RGB enables standard RGB-based perception models to operate robustly on LiDAR-only inputs across multiple downstream tasks and diverse lighting conditions. The source code and the supplementary video are available at: \href{https://github.com/IMRL/Cyclops}{https://github.com/IMRL/Cyclops}.
    
\end{abstract}



\section{Introduction}
\label{sec:intro}
\begin{figure}
    \centering
    \includegraphics[width=0.85\linewidth]{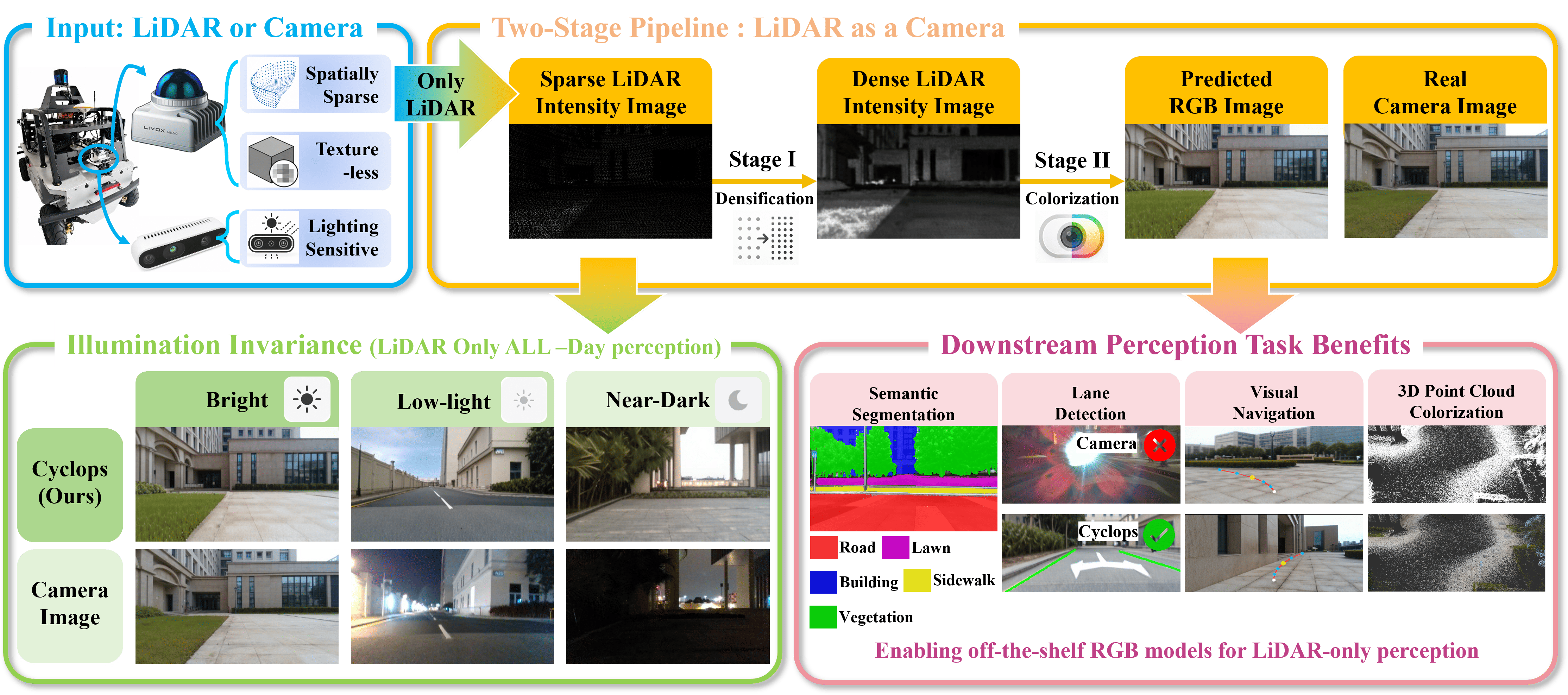}
    \caption{Our proposed \textit{Cyclops} converts sparse NRS-LiDAR intensity into RGB imagery through a two-stage pipeline: densification (Stage~I) followed by colorization (Stage~II). \textbf{Bottom-left:} Generated images maintain consistent visual quality across diverse lighting conditions, whereas camera images degrade severely. \textbf{Bottom-right:} The synthesized RGB directly enables downstream tasks, including semantic segmentation, lane detection, visual navigation, and point cloud colorization.}
    \label{fig:cover}
    \vspace{-20pt}
\end{figure}
Robust perception is a fundamental prerequisite for autonomous mobile robots~\citep{cadena2016past}. Among onboard sensors, cameras provide dense, high-resolution imagery rich in color and semantic cues, but are inherently sensitive to illumination variations that compromise perception reliability~\citep{gao2025night,gao2026super}. In contrast, LiDAR actively captures precise 3D geometry along with intensity, which encodes illumination-invariant reflectance information with minimal dependence on ambient lighting.

Recent studies have demonstrated the value of LiDAR intensity in tasks such as loop closure detection~\citep{shan2021robust,di2021visual}, odometry~\citep{pfreundschuh2024coin,zhang2023ri}, and segmentation~\citep{gao2024active,viswanath2024reflectivity}. However, these methods typically rely on expensive high-resolution LiDAR, and the single-channel nature of intensity fundamentally limits its perceptual capacity compared to RGB imagery. Existing approaches that attempt to bridge this gap, including GAN-based LiDAR-to-camera translation~\citep{cortinhal2021semantics,xu2025ligencam}, neural rendering~\citep{xu2022point,huang2023neural}, and diffusion-based synthesis~\citep{gao2023magicdrive,wen2024panacea}, either require dense inputs, per-scene optimization with RGB supervision, or auxiliary annotations with slow iterative inference, limiting practical applicability.

In this work, we aim to achieve image-level scene perception using only a low-cost NRS-LiDAR, with no camera input or auxiliary annotations required at inference. This goal poses three main challenges: \textbf{(i) Data Sparsity:} NRS-LiDAR yields sparse, non-uniformly sampled projections with large missing regions; \textbf{(ii) Structural Fidelity under Modality Gap:} the single-channel intensity differs significantly from RGB in distribution, requiring cross-modal translation that preserves structural consistency; \textbf{(iii) Temporal Stability and Deployment Efficiency:} the method should mitigate inter-frame flickering and support efficient inference with few network evaluations.

To address these challenges, we present \textit{Cyclops}, a unified framework that transforms sparse NRS-LiDAR scans into RGB video suitable for robotic perception (see Fig.~\ref{fig:cover}). For (i), we adopt the densification approach from~\citep{gao2026super} as a frozen geometric prior (Stage~I). For (ii), we formulate intensity-image-to-RGB translation as a latent-space transport problem within the LBM~\citep{chadebec2025lbm} (Stage~II); since both modalities share scene geometry, the latent displacement primarily encodes appearance, enabling accurate generation in only a few Euler steps. For (iii), we augment the velocity field with temporal attention for inter-frame context and further treat it as a policy optimized by a differentiable terminal reward backpropagated through the ODE trajectory using Backpropagation Through Time. The main contributions are:
(1) To our knowledge, this is the first work to synthesize RGB imagery from sparse NRS-LiDAR observations alone, effectively converting a low-cost NRS-LiDAR into an illumination-invariant ``All-Day Camera''.
(2) We present an end-to-end, perception-oriented system that integrates a frozen LiDAR intensity densification module with latent bridge matching, temporal conditioning, and reward-guided trajectory optimization, balancing generation quality and inference efficiency.
(3) We validate the synthesized RGB on semantic segmentation, lane detection, pretrained ViNT~\citep{shah2023vint} navigation, and point cloud colorization, demonstrating substantial improvements over both LiDAR baselines and cameras under diverse lighting conditions.
(4) We release the source code at: \href{https://github.com/IMRL/Cyclops}{https://github.com/IMRL/Cyclops}.
\section{Related Works}
\label{sec: 2}

\subsection{LiDAR Intensity in Robotic Perception}
\label{subsec:intensity_densification}

LiDAR intensity, representing the surface reflectivity of objects, provides crucial textural information that complements pure geometric sensing. Intensity data has been combined with geometric cues to enhance loop closure detection, yielding more robust place recognition in large-scale environments~\citep{shan2021robust, wang2020lidar, di2021visual} as well as reliable terrain and material classification~\citep{gao2024active, viswanath2024reflectivity}. Furthermore, intensity-enhanced odometry frameworks have integrated intensity images with point cloud registration to maintain localization where geometry alone is insufficient~\citep{pfreundschuh2024coin, zhang2023ri, du2023real}.

Despite the demonstrated utility of intensity as a stable texture source, existing methods face two fundamental limitations: dependence on expensive, high-resolution LiDAR for sufficiently dense measurements, and the inherently restricted perceptual capacity of single-channel data relative to RGB. These limitations motivate our approach of first densifying sparse intensity into high-resolution images and then translating them into RGB, bridging the gap between low-cost LiDAR and camera-level perception within a unified framework.

\subsection{Cross-Modal Generation for Robotic Perception}
\label{subsec:related_crossmodal}

Converting LiDAR intensity images into RGB can be formulated as cross-modal image-to-image translation, requiring faithful preservation of structural layout while synthesizing realistic colors and textures. Early efforts predominantly adopted Generative Adversarial Networks (GANs). Pix2Pix~\citep{isola2017image} established paired conditional translation, while CycleGAN~\citep{zhu2017unpaired} relaxed this to unpaired settings via cycle-consistency. These were extended to LiDAR-to-camera tasks: Cortinhal~\textit{et al.}~\citep{cortinhal2021semantics,cortinhal2024depth} translated LiDAR scans into panoramic RGB-D images, LiGenCam~\citep{xu2025ligencam} reconstructed camera views from multimodal LiDAR inputs, and Ha~\textit{et al.}~\citep{ha2025enhancing} applied colorization and super-resolution to LiDAR imagery for odometry. Despite offering fast inference, GAN-based methods generally suffer from texture hallucinations that undermine physical plausibility~\citep{park2019semantic,wang2018high}.

Denoising Diffusion Probabilistic Models (DDPMs)~\citep{ho2020denoising,sohl2015deep} surpass GANs in sample quality and mode coverage through iterative denoising. LDM~\citep{rombach2022high} reduces cost by operating in latent space, while ControlNet~\citep{zhang2023adding} and T2I-Adapter~\citep{mou2024t2i} enable spatially guided generation from structural inputs. For cross-modal translation, DCLTV~\citep{zhang2025dcltv} converts active laser images to visible light, and Veila~\citep{liu2025veila} synthesizes LiDAR scenes from monocular RGB. However, iterative sampling poses a critical latency bottleneck. Acceleration methods such as DDIM~\citep{song2020denoising}, Consistency Models~\citep{song2023consistency}, and LCM~\citep{luo2023latent} reduce step counts but often at the cost of blurred details and structural coherence.

Video-to-video synthesis methods offer an alternative paradigm for temporally consistent generation~\citep{wang2018video,blattmann2023align,yang2025cogvideox}. However, these approaches are ill-suited to robotic deployment: most architectures are non-causal, requiring future frames for coherent generation and thus violating online streaming constraints, while temporal 3D attention combined with multi-step denoising yields per-frame latency on the order of seconds~\citep{ho2022video}, far exceeding real-time requirements. To reconcile fidelity with efficiency, Rectified Flow~\citep{liu2022flow} and Schr\"{o}dinger Bridges~\citep{liu20232,graikos2022diffusion} learn direct trajectories between source and target distributions rather than mapping from Gaussian noise. Building on this paradigm, our method adopts LBM to establish a deterministic intensity-image-to-RGB mapping. We further integrate temporal attention for inter-frame context and a differentiable reward-guided trajectory optimization to encourage temporal stability and semantic fidelity--properties that are important for robotic perception but are not explicitly addressed in most prior generative approaches.

\section{Methodology}
\label{sec:methodology}

\subsection{Problem Formulation and Pipeline Overview}
\label{subsec:overview}

\begin{figure}[t]
    \centering
    \includegraphics[width=0.75\linewidth]{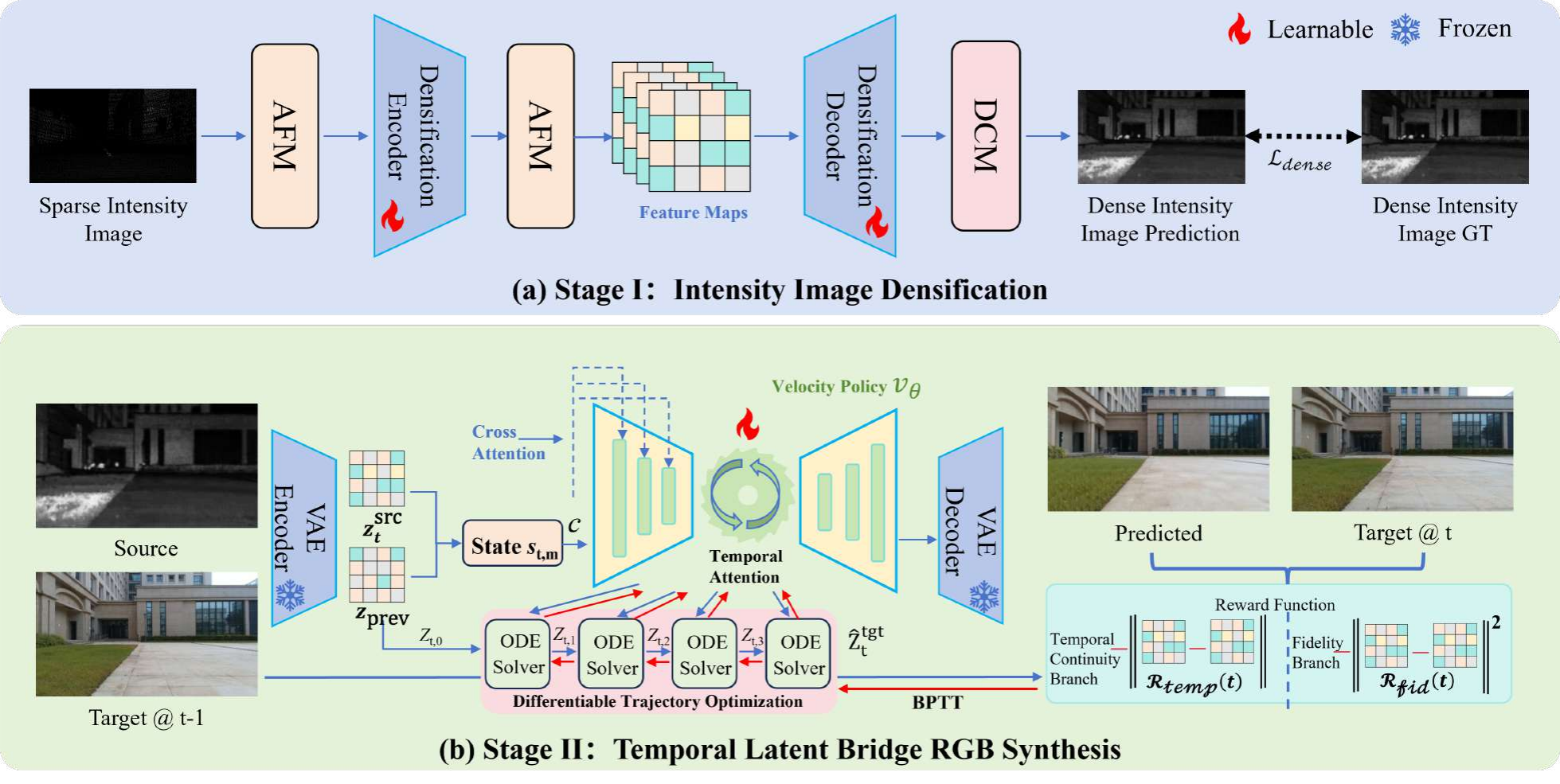}
    \caption{Overview of our proposed framework. \textbf{(a)~Stage~I} densifies LiDAR intensity image via densification network with AFM and DCM. \textbf{(b)~Stage~II} transports the dense intensity image latent toward an RGB latent via a 4-step ODE governed by velocity policy $v_\theta$, with temporal attention for inter-frame conditioning and a differentiable reward $\mathcal{R}$ for end-to-end trajectory optimization.}
    \label{fig:framework}
    \vspace{-20pt}
\end{figure}

Given a temporal sequence of sparse LiDAR intensity scans 
$\{x^{\mathrm{sparse}}_t\}_{t=1}^{T}$, our goal is to synthesize a corresponding sequence of RGB images 
$\{\hat{y}_t\}_{t=1}^{T}$ that are structurally aligned with the 3D scene geometry, semantically faithful to the ground-truth appearance, and temporally coherent across consecutive frames. As illustrated in Fig.~\ref{fig:framework}, the pipeline consists of two stages. In \textbf{Stage~I} (Sec.~\ref{subsec:stage1}), a frozen pre-trained densification network $\mathcal{F}_{\mathrm{dense}}$ converts each sparse input $x^{\mathrm{sparse}}_t$ into a dense intensity image $I^{\mathrm{dense}}_t$. In \textbf{Stage~II} (Sec.~\ref{subsec:stage2}), $I^{\mathrm{dense}}_t$ is encoded into a source latent $z^{\mathrm{src}}_t=\mathcal{E}(I^{\mathrm{dense}}_t)$ via a pre-trained VAE encoder, and the temporal latent bridge module learns a transport trajectory from $z^{\mathrm{src}}_t$ toward the ground-truth RGB latent $z^{\mathrm{gt}}_t=\mathcal{E}(y_t)$. The final RGB image is decoded as $\hat{y}_t=\mathcal{D}(\hat{z}^{\mathrm{tgt}}_t)$. Throughout this paper, subscript $t$ denotes the frame index and $\tau\in[0,1]$ denotes the continuous time within the ODE-based generative process of a single frame.

\textbf{Supervision and Data Pairing.} Our dataset combines self-collected sequences and the public dataset M3DGR ~\citep{zhang2025towards}. Raw LiDAR intensities are projected onto the camera plane via pre-calibrated extrinsics, yielding sparse intensity images pixel-aligned and time-synchronized with RGB targets. For training, we curate sequences with minimal dynamic objects to ensure clean supervision; for testing, the moving-object filter of~\citep{wu2024moving} is applied prior to multi-scan fusion to remove motion artifacts. The train/validation/test sets are split by scene, with no geographic overlap.  

\subsection{Stage I: Intensity Image Densification}
\label{subsec:stage1}

\begin{wrapfigure}{r}{0.45\textwidth}
    \centering
    \includegraphics[width=0.9\linewidth]{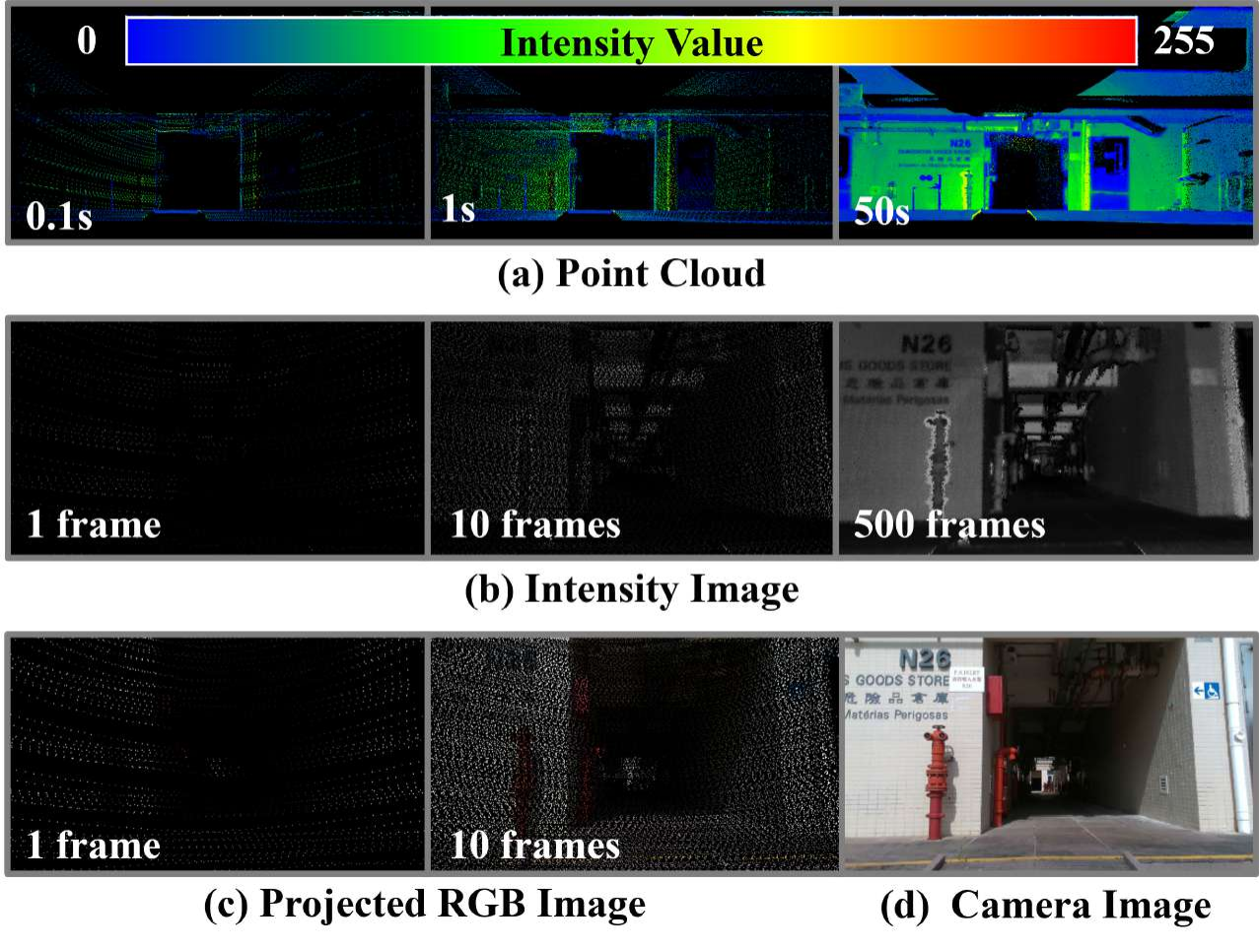}
    \caption{NRS-LiDAR scan accumulation and sparsity challenge. (a)~Point cloud progressively accumulates. (b)~Accumulated intensity map projected onto the camera viewpoint. (c)~LiDAR points projected onto the image plane and colorized with corresponding camera RGB values. (d)~Reference camera image. Comparing (c) and (d) reveals significant texture loss and large spatial gaps in short-duration observations.}
    \label{fig:accumulate}
    \vspace{-20pt}
\end{wrapfigure}

As shown in Fig.~\ref{fig:accumulate}, directly projecting short-duration LiDAR scans yields severe sparsity that makes direct intensity-to-RGB translation ill-posed. Fortunately, NRS-LiDARs steer the beam along non-repeating trajectories, enabling progressive coverage over time~\citep{gao2026super}. This property allows constructing paired sparse--dense training data from stationary accumulation. While Cyclops is developed and trained on Livox MID-360 data, we further extend Stage~I with a training-free cross-LiDAR adaptation that aligns the projected field of view, sampling sparsity, and intensity statistics of other sensors with the original Livox input domain; details are provided in Appendix~\ref{app:generalization}.
\vspace{-7pt}

We leverage the densification framework from~\citep{gao2026super}, which adopts a U-shaped encoder-decoder architecture augmented with two task-specific modules: (i)~an \textit{Adaptive Fusion Module}~(AFM) that employs parallel dilated and deformable convolutions to aggregate multi-scale context, recovering structure in large void regions while preserving edges; and (ii)~a \textit{Dynamic Compensation Module}~(DCM) that calibrates the decoder output using geometric cues such as range and incidence angle to produce physically consistent intensity values (details in Appendix~\ref{app:stage1_arch}). The network is supervised by MSE reconstruction loss and trained on the \textit{Super LiDAR Intensity} dataset~\citep{gao2026super}, which provides paired sparse--dense intensity images.

In this work, the pre-trained $\mathcal{F}_{\mathrm{dense}}$ is used as a frozen module. Given $N{=}5$ consecutive LiDAR scans spatiotemporally aligned via the generic and versatile LiDAR-inertial odometry framework~\citep{zhao2024fmcw} and projected onto a 2D image plane, it produces a dense intensity image $I^{\mathrm{dense}}_t=\mathcal{F}_{\mathrm{dense}}(\mathbf{X}_{\mathrm{sparse},t})$ that captures geometry-consistent texture information. This dense image is then encoded as $z^{\mathrm{src}}_t=\mathcal{E}(I^{\mathrm{dense}}_t)$ to initialize Stage~II.

\subsection{Stage II: Temporal Latent Bridge RGB Synthesis}
\label{subsec:stage2}

\subsubsection{Preliminaries: Latent Bridge Matching}
\label{subsec:lbm}

Our generative backbone builds upon LBM~\citep{chadebec2025lbm}, which learns a
direct transport map between two distributions in latent space. Given the source
latent $z^{\mathrm{src}}\sim\pi_0$ and the ground-truth RGB latent
$z^{\mathrm{gt}}\sim\pi_1$, an intermediate bridge sample is constructed as
$z_\tau = (1{-}\tau)\,z^{\mathrm{src}} + \tau\,z^{\mathrm{gt}}
+ \sigma\sqrt{\tau(1{-}\tau)}\,\epsilon$ with
$\epsilon\sim\mathcal{N}(0,I)$ and $\sigma\geq 0$ controlling stochasticity.
A velocity field $v_\theta(z_\tau,\tau,c)$ is trained to regress the conditional
drift of the bridge process:
\begin{equation}
    \mathcal{L}_{\mathrm{LBM}} =
    \mathbb{E}_{z^{\mathrm{src}},\,z^{\mathrm{gt}},\,\tau}
    \left[
    \left\|
    v_\theta(z_\tau, \tau, c) -
    \frac{z^{\mathrm{gt}} - z_\tau}{1 - \tau}
    \right\|_2^2
    \right],
    \label{eq:lbm_loss}
\end{equation}
where $\tau\in[0,1]$ denotes the bridge time. Following~\citep{chadebec2025lbm}, we restrict $\tau$ during training to only $M{=}4$ equally spaced values that coincide with the inference discretization, acting as implicit distillation and enabling high-quality generation with only 4 Euler steps at test time.



\subsubsection{Temporal Conditioning Mechanism}
\label{subsec:temporal_cond}

Independently generating each frame may introduce inter-frame flickering and geometric inconsistency, which can degrade sequential robotic perception. We therefore condition the velocity field at frame $t$ and ODE step $m$ on the state
$s_{t,m}=(z_{t,m},z^{\mathrm{src}}_t,z_{\mathrm{prev}})$.
At each U-Net resolution, let
$\mathcal{A}_{W}(a,b)=\mathrm{Attn}(W_q a,W_k b,W_v b)$,
where the first input provides the query and the second provides the key and value. Source and temporal conditioning are then applied sequentially:
\begin{equation}
\begin{aligned}
h'  &= h+\mathcal{A}_{W}\!\left(h,z^{\mathrm{src}}_t\right),\\
h'' &= h'+\mathcal{A}_{W'}\!\left(h',z_{\mathrm{prev}}\right),
\end{aligned}
\label{eq:temporal_conditioning}
\end{equation}
where $W$ and $W'$ denote independent learnable projections. The source branch anchors generation to the current LiDAR intensity geometry, while the temporal branch retrieves appearance cues from the previous target latent. During teacher-forced training, $z_{\mathrm{prev}}=z^{\mathrm{gt}}_{t-1}$; during model-conditioned training and inference, $z_{\mathrm{prev}}=\hat{z}^{\mathrm{tgt}}_{t-1}$, and a learnable null token $z_{\varnothing}$ is used for the first frame. Because the temporal branch acts as an appearance memory rather than performing pixel-wise fusion, it does not require exact spatial alignment between consecutive frames and reduces flickering without overriding the current-frame geometry (details are provided in Appendix~\ref{app:temporal_attn}).

\subsubsection{Differentiable Reward-Guided Trajectory Optimization}
\label{subsec:temporal_opt}

The bridge matching loss (Eq.~\ref{eq:lbm_loss}) trains the velocity at arbitrary intermediate states but does not explicitly optimize the terminal output of the multi-step ODE for task-specific objectives. We address this by treating the velocity field as a \textbf{velocity policy} and introducing a differentiable terminal reward backpropagated through the ODE trajectory~\citep{clark2024directly}.

\textbf{Velocity Policy.}
At the $m$-th Euler step, the policy predicts:
\begin{equation}
    z_{t,m+1} =
    z_{t,m} +
    v_\theta\!\left(
    z_{t,m},\,\tfrac{m}{M},\,c_t
    \right)\Delta\tau,
    \quad m = 0,\ldots,M{-}1,
    \label{eq:ode_steps_policy}
\end{equation}
where $\Delta\tau=1/M$ and $c_t$ encodes both source and temporal conditioning. The terminal latent $\hat{z}^{\mathrm{tgt}}_t=z_{t,M}$ is decoded to produce $\hat{y}_t$.

\textbf{Reward Design.}
The terminal reward combines spatial fidelity with temporal coherence:
\begin{equation}
    \mathcal{R}(t) =
    \omega_1 \mathcal{R}_{\mathrm{fid}}(t) +
    \omega_2 \mathcal{R}_{\mathrm{temp}}(t),
    \label{eq:reward_total}
\end{equation}
where $\mathcal{R}_{\mathrm{fid}}(t)=-\|\hat{z}^{\mathrm{tgt}}_t-z^{\mathrm{gt}}_t\|_2^2$ encourages fidelity, and the temporal continuity reward penalizes abnormal temporal displacement:
\begin{equation}
    \mathcal{R}_{\mathrm{temp}}(t) =
    -\left\|
    \left(\hat{z}^{\mathrm{tgt}}_t - z_{\mathrm{prev}}\right)
    -
    \left(z^{\mathrm{gt}}_t - z^{\mathrm{gt}}_{t-1}\right)
    \right\|_1.
    \label{eq:reward_temp}
\end{equation}
This formulation preserves real scene motion while suppressing flickering, by matching generated temporal displacement to the ground-truth displacement rather than directly minimizing inter-frame differences (further discussion in Appendix~\ref{app:reward_extended}).

\textbf{Optimization via Backpropagation Through Time (BPTT).}
Since the Euler integration (Eq.~\ref{eq:ode_steps_policy}) is fully differentiable, the terminal reward gradient with respect to $\theta$ is computed by backpropagation through the $M{=}4$ steps, propagating the reward signal to every intermediate velocity prediction. This avoids the high variance of stochastic policy gradient estimators and enables stable optimization of task-level objectives through the generation dynamics.

\subsection{Training Strategy}
\label{subsec:training}

Training proceeds in two phases. In \textbf{Phase~1}, the velocity field is trained with teacher-forced temporal conditioning ($z_{\mathrm{prev}}=z^{\mathrm{gt}}_{t-1}$) using a composite objective: $\mathcal{L}_{\mathrm{Phase1}}=\mathcal{L}_{\mathrm{LBM}} +\lambda_1\mathcal{L}_{\mathrm{lpips}}
+\lambda_2\mathcal{L}_{\mathrm{grad}}
+\lambda_3\mathcal{L}_{\mathrm{color}}$,
where $\mathcal{L}_{\mathrm{lpips}}$ is the LPIPS perceptual 
distance~\citep{zhang2018unreasonable}, $\mathcal{L}_{\mathrm{grad}}$ enforces edge sharpness via Sobel gradient matching, and $\mathcal{L}_{\mathrm{color}}$ constrains channel-wise color statistics (detailed definitions in Appendix~\ref{app:training_details}). In \textbf{Phase~2}, we augment the objective with the differentiable 
terminal reward and adopt scheduled sampling, linearly annealing 
teacher-forcing probability from 1.0 to 0.2 to mitigate exposure bias:
\begin{equation}
    \mathcal{L}_{\mathrm{Phase2}} =
    \mathcal{L}_{\mathrm{Phase1}} -
    \lambda_4\,\mathcal{R}(t).
    \label{eq:phase2_loss}
\end{equation}
The learning rate is reduced by $10\times$ to preserve the generative prior learned in Phase~1. All hyperparameters are detailed in Appendix~\ref{app:training_details}.
\section{Experiments}
\label{sec:experiments}

We evaluate the proposed method from two perspectives: the intrinsic quality of LiDAR intensity image colorization (\S\ref{subsec:colorization_eval}), and its practical value in four downstream perception tasks-semantic segmentation (\S\ref{subsec:seg}), traffic lane detection (\S\ref{subsec:lane}), visual navigation (\S\ref{subsec:navigation}), and point cloud colorization (\S\ref{subsec:pc_color}). All data are collected using a Livox MID-360 mounted on a ground robot equipped with an Intel i7-1165G7 CPU. Our dataset contains over 40 sequences totaling more than 28,000 synchronized sparse intensity image-RGB pairs, all at $256{\times}455$ resolution, split into training/validation/testing at a 6:1:1 ratio (dataset detail in Appendix~\ref{app:dataset}). Following LBM~\citep{chadebec2025lbm}, we use the frozen pre-trained VAE from SDXL~\citep{podell2024sdxl} and train Stage~II on 2 H800 GPUs. Experiments are conducted under three lighting conditions:
Bright (sufficient natural illumination), Low Light (insufficient natural light with artificial sources), and Near Dark (near-total absence of illumination).

\subsection{Evaluation of Temporal Latent Bridge RGB Synthesis}
\label{subsec:colorization_eval}

\subsubsection{Comparison with Baselines}

\begin{figure*}[t]
    \centering
    \includegraphics[width=0.85\linewidth]{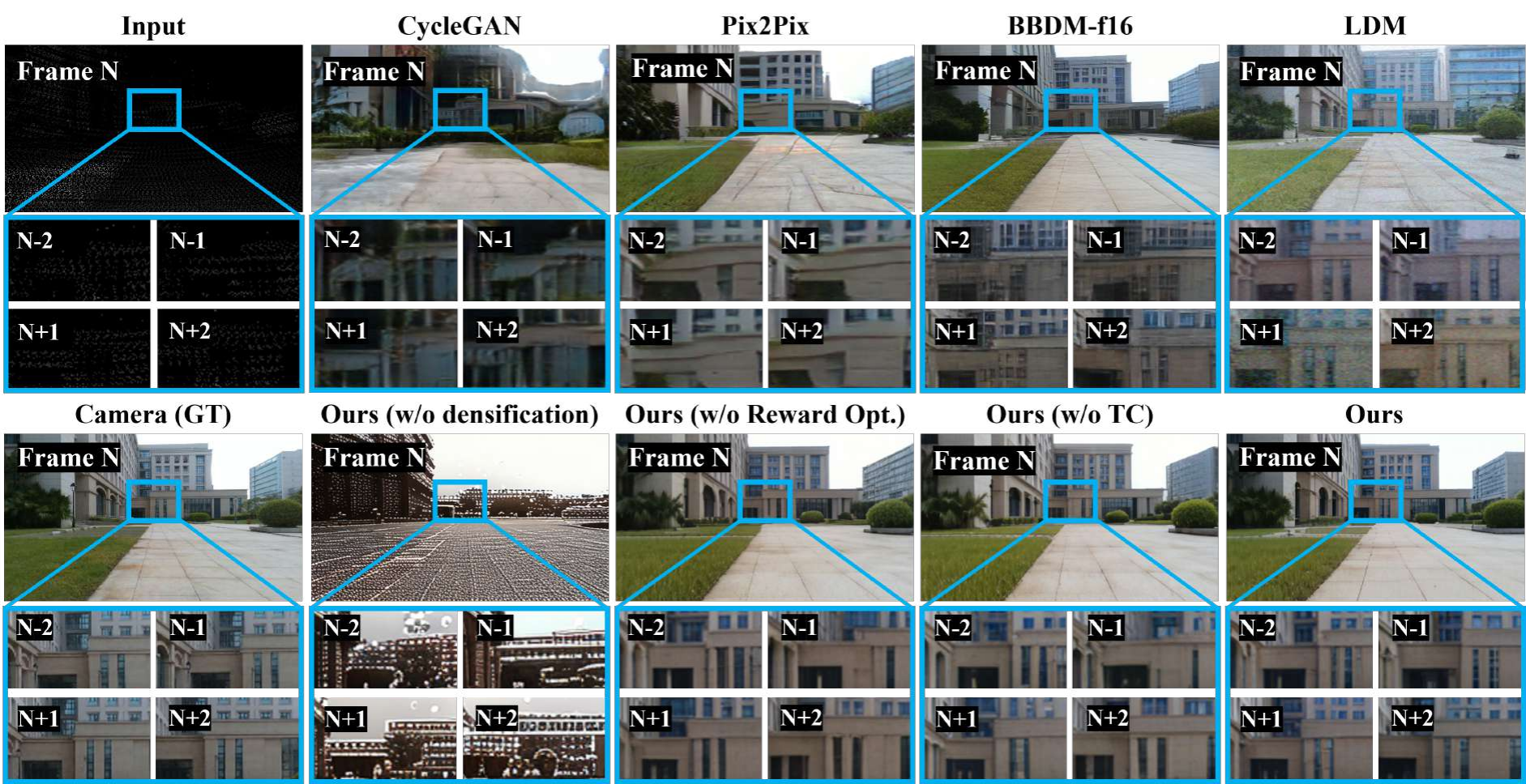}
    \caption{Qualitative comparison. The top row shows baseline results; the bottom row presents ablation variants. Each group includes the full image at frame $N$ and zoomed-in patches from frames $N{-}2$ to $N{+}2$ to visualize temporal consistency. Our full model produces the most realistic colors with faithful geometry and strong temporal coherence.}
    \label{fig:qualitative_results}
    \vspace{-15pt}
\end{figure*}

We compare against GAN-based methods (CycleGAN~\citep{zhu2017unpaired}, Pix2Pix~\citep{isola2017image}), diffusion/bridge-based methods (BBDM~\citep{li2023bbdm}, LDM~\citep{rombach2022high}), and recent controllable generators (CtrLoRA~\citep{xu2025ctrlora}, OminiControl~\citep{tan2025ominicontrol}).
All baselines and ablation variants receive the same Stage~I densified intensity images as input, ensuring fair comparison; CtrLoRA and OminiControl are trained on the same paired data and conditioning input as Cyclops. The runtime reported in Table~\ref{tab:colorization} measures Stage~II only, while Stage~I adds 0.043-0.055 s per frame (measured on an NVIDIA RTX 3090 GPU). OminiControl exceeds 24~GB memory at full resolution, so its runtime is reported as out of memory (OOM). We evaluate image translation quality via PSNR, SSIM, LPIPS~\citep{zhang2018unreasonable}, and FID~\citep{heusel2017gans}, and temporal consistency via Mean Frame Difference (MFD), Flow-Consistency MSE (FC-MSE), and Flicker Variance (FV). As shown in Table~\ref{tab:colorization} and Fig.~\ref{fig:qualitative_results}, GAN-based methods yield the lowest quality and worst temporal stability. Diffusion-based and controllable methods improve perceptual quality, but remain less temporally consistent and substantially slower than Cyclops (detailed in Appendix~\ref{app:baseline_analysis}).
\begin{table}
    \centering
    \caption{Quantitative comparison of image translation quality and temporal consistency. The upper block lists baseline methods; the lower block presents our ablation study.}
    \label{tab:colorization}
    \resizebox{\columnwidth}{!}{
    \begin{tabular}{l cc cc cc cc c}
        \toprule
        \multirow{2}{*}{\textbf{Method}} &
        \multicolumn{4}{c}{\textbf{Image Translation Quality}} &
        \multicolumn{3}{c}{\textbf{Temporal Consistency}} &
        \multirow{2}{*}{\textbf{Time (s)}$\downarrow$} \\
        \cmidrule(lr){2-5} \cmidrule(lr){6-8}
        & \textbf{PSNR}$\uparrow$ & \textbf{SSIM}$\uparrow$
        & \textbf{LPIPS}$\downarrow$ & \textbf{FID}$\downarrow$
        & \textbf{MFD}$\downarrow$ & \textbf{FC-MSE}$\downarrow$
        & \textbf{FV}$\downarrow$ & \\
        \midrule
        CycleGAN~\citep{zhu2017unpaired}
            & 14.25 & 0.452 & 0.385 & 138.5
            & 38.56 & 0.0478 & 22.35 & 0.028 \\
        Pix2Pix~\citep{isola2017image}
            & 16.34 & 0.526 & 0.318 & 102.3
            & 31.82 & 0.0385 & 17.62 & 0.022 \\
        BBDM~\citep{li2023bbdm}
            & 20.12 & 0.625 & 0.238 & 68.5
            & 26.48 & 0.0312 & 14.56 & 2.75 \\
        LDM~\citep{rombach2022high}
            & 19.58 & 0.608 & 0.252 & 74.2
            & 33.25 & 0.0415 & 18.92 & 3.28 \\
        CtrLoRA~\citep{xu2025ctrlora}
            & 17.47 & 0.553 & 0.307 & 90.7
            & 41.44 & 0.0521 & 27.01 & 13.6 \\
        OminiControl~\citep{tan2025ominicontrol}
            & 18.02 & 0.586 & 0.260 & 82.4
            & 35.36 & 0.0458 & 21.19 & OOM \\
        \midrule
        Ours w/o Densification
            & 11.23 & 0.312 & 0.528 & 185.6
            & 46.72 & 0.0685 & 30.15 & 0.235 \\
        Ours w/o Reward Opt.
            & 21.85 & 0.662 & 0.205 & 53.8
            & \underline{15.38} & \underline{0.0162} & \underline{7.85} & 0.212 \\
        Ours w/o TC
            & \underline{22.12} & \underline{0.675} & \underline{0.192} & \underline{48.6}
            & 22.65 & 0.0272 & 12.48 & 0.193 \\
        \textbf{Full (Ours)}
            & \textbf{22.48} & \textbf{0.692} & \textbf{0.178} & \textbf{43.5}
            & \textbf{14.25} & \textbf{0.0148} & \textbf{7.12} & 0.238 \\
        \bottomrule
    \end{tabular}
    }
    \vspace{-10pt}
\end{table}

\subsubsection{Ablation Study}
\label{ablation}

The lower block of Table~\ref{tab:colorization} isolates the contribution of each component. Removing Stage~I densification causes the most severe degradation, confirming it as a fundamental prerequisite for tractable dense-to-dense translation. Removing reward-guided mainly impacts perceptual quality while temporal consistency remains largely intact, indicating that the reconstruction-only objective lacks the structural precision provided by reward-driven optimization. Removing the temporal consistency module preserves per-frame quality but nearly doubles MFD, confirming that high single-frame fidelity alone does not guarantee stable sequential outputs. The full model achieves the best performance across all metrics, confirming the three components are complementary.

\subsection{Semantic Segmentation}
\label{subsec:seg}

We evaluate semantic segmentation on over 1,000 test frames with manually annotated pixel-level masks covering five classes: Road, Sidewalk, Vegetation, Building, and Lawn. For all image-based inputs--camera images, densified intensity images, and Cyclops-rendered RGB--we use the same RGB-pretrained SAM\,2~\citep{ravi2025sam} model without task-specific fine-tuning. SAM\,2 is operated in video-predictor mode, where identical point prompts are provided across all three modalities every five frames, with 3-8 representative points for each semantic class present in a prompted frame. Masks for the intervening unprompted frames are propagated through its temporal memory. As LiDAR-specific references, we evaluate Patchwork++~\citep{lee2022patchwork++} and GroundGrid~\citep{steinke2023groundgrid}. We also considered FRNet and RangeViT as LiDAR segmentation baselines; however, their range/frustum representations do not naturally accommodate the MID-360's irregular, time-varying scan geometry. 

\begin{table*}[t]
    \centering
    \caption{Semantic segmentation under varying illumination. SAM\,2 is evaluated in video-predictor mode using identical point prompts every five frames for all image modalities. $^\dagger$ denotes binary ground/non-ground segmentation methods whose outputs are mapped to the evaluation labels as described in Appendix~\ref{app:seg_setup}.}
    \label{tab:segmentation}
    \resizebox{\textwidth}{!}{
    \begin{tabular}{ll ccc ccc ccc}
        \toprule
        \multirow{2}{*}{\textbf{Method}} &
        \multirow{2}{*}{\textbf{Input}} &
        \multicolumn{3}{c}{\textbf{Bright}} &
        \multicolumn{3}{c}{\textbf{Low Light}} &
        \multicolumn{3}{c}{\textbf{Near Dark}} \\
        \cmidrule(lr){3-5} \cmidrule(lr){6-8} \cmidrule(lr){9-11}
        & & mIoU & mAcc & aAcc
          & mIoU & mAcc & aAcc
          & mIoU & mAcc & aAcc \\
        \midrule
        Patchwork++$^\dagger$~\citep{lee2022patchwork++}
            & Point Cloud
            & 44.2 & 53.5 & 73.6
            & 44.0 & 53.2 & 73.4
            & 43.6 & 52.8 & 73.1 \\
        GroundGrid$^\dagger$~\citep{steinke2023groundgrid}
            & Point Cloud
            & 47.5 & 56.8 & 75.8
            & 47.2 & 56.5 & 75.5
            & 46.8 & 56.1 & 75.2 \\
        \midrule
        \multirow{3}{*}{SAM\,2~\citep{ravi2025sam}}
            & Densified Intensity
            & 35.2 & 43.8 & 67.2
            & 35.0 & 43.5 & 67.0
            & 34.7 & 43.2 & 66.7 \\
            & Camera
            & \textbf{76.8} & \textbf{84.5} & \textbf{92.8}
            & 63.5 & 72.2 & \textbf{86.8}
            & 31.2 & 39.8 & 63.2 \\
            & Cyclops Rendering
            & 65.8 & 75.2 & 86.2
            & \textbf{65.2} & \textbf{74.5} & 85.6
            & \textbf{64.5} & \textbf{73.8} & \textbf{85.1} \\
        \bottomrule
    \end{tabular}}
    \vspace{-5pt}
\end{table*}

As shown in Table~\ref{tab:segmentation}, applying SAM\,2 directly to densified intensity images yields only 34.7-35.2 mIoU, whereas Cyclops improves the results to 64.5-65.8 mIoU using the same model and prompting protocol. This approximately 30-point improvement indicates that the geometry-conditioned RGB rendering provides semantic cues that are substantially more compatible with RGB-pretrained representations than raw intensity inputs. Camera-based segmentation achieves the best result under Bright conditions but decreases from 76.8 to 31.2 mIoU under Near Dark. In contrast, Cyclops varies by only 1.3 mIoU points across the three illumination conditions.
\begin{figure*}[t]
    \centering
    \includegraphics[width=0.9\textwidth]{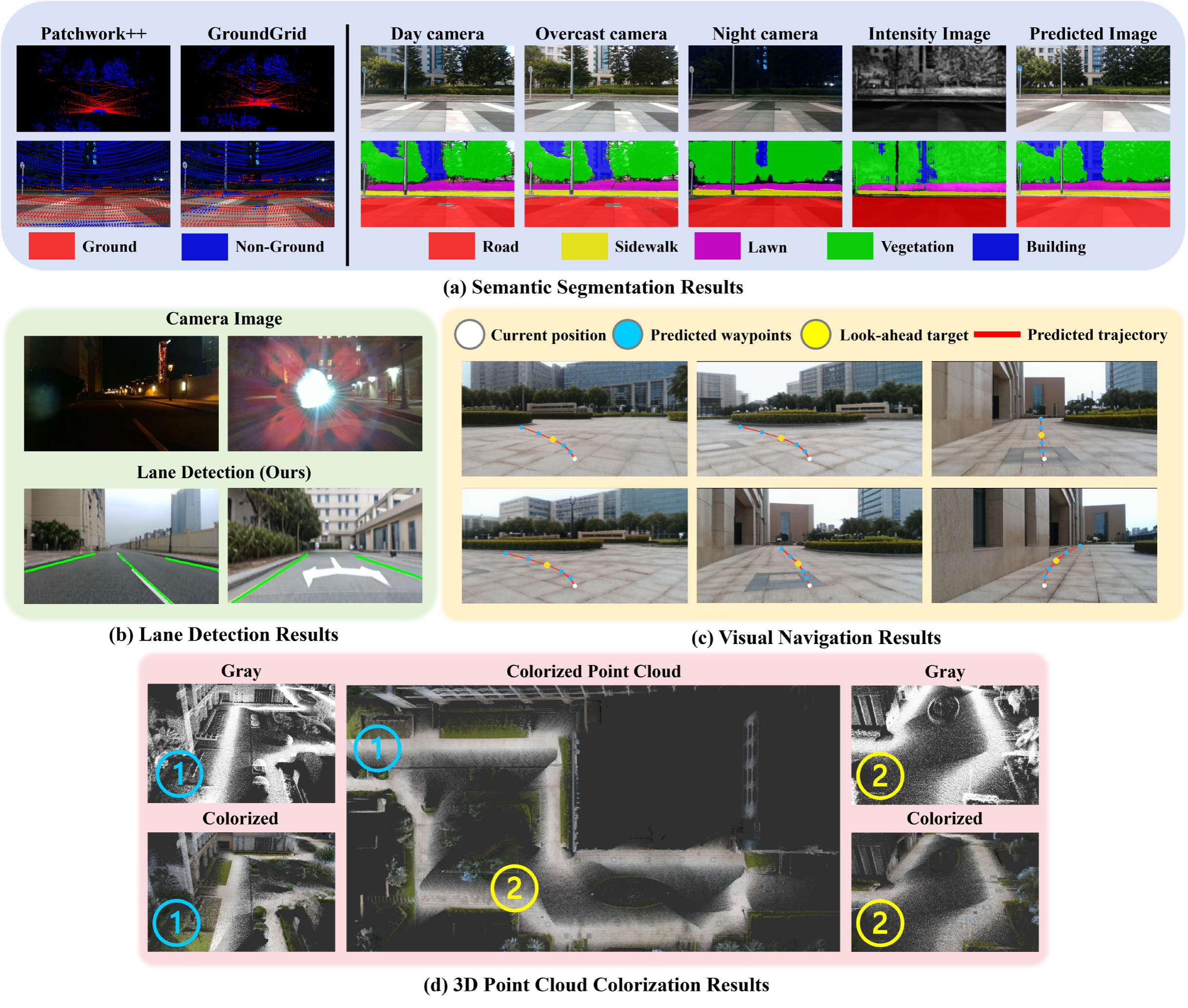}
    \caption{Qualitative results of downstream tasks.
    \textbf{(a)} Left: ground segmentation from Patchwork++ and GroundGrid projected onto the image plane. Right: semantic segmentation across different inputs--camera-based results degrade under poor lighting, whereas our colorized images maintain consistent quality.
    \textbf{(b)} Lane detection on Cyclops-generated RGB images under challenging lighting conditions.
    \textbf{(c)} Visual navigation using pretrained ViNT, which directly takes Cyclops-generated RGB images as input to predict waypoint trajectories.
    \textbf{(d)} Compared with grayscale point clouds, our colorized point clouds exhibit richer texture details.}
    \label{fig:task}
    \vspace{-15pt}
\end{figure*}
\subsection{Traffic Lane Detection}
\label{subsec:lane}

We adopt LaneATT~\citep{tabelini2021keep} with a ResNet-34 backbone~\citep{he2016deep}, pre-trained on TuSimple~\citep{tusimple-lane-detection} and fine-tuned on our domain ($\sim$1,300 training images). Three copies of the same pre-trained model are fine-tuned identically: one on camera images, one on densified intensity images, and one on our colorized images. Full experimental setup and quantitative results are provided in Appendix~\ref{app:lane_details}.

The intensity-only model achieves reasonable accuracy (93.3\%-96.1\%) across all lighting conditions, confirming that lane markings inherently produce strong reflectance contrast regardless of illumination. \textit{Cyclops} further improves performance (accuracy 95.2\%-95.8\%), as RGB texture features help resolve ambiguous cases such as worn markings or shadow boundaries. Under Bright conditions, the camera-based model achieves the highest accuracy (96.5\%); however, it degrades severely under Near Dark (accuracy 71.2\%), whereas both intensity-only and colorized models maintain stable performance (accuracy $>$95\%). This confirms that material reflectance is the dominant cue and that colorization provides complementary texture features for borderline cases (see Fig.~\ref{fig:task}(b)).


\subsection{Visual Navigation}
\label{subsec:navigation}

We further evaluate Cyclops in a navigation task across different illumination conditions. Under identical settings, either Cyclops-generated RGB or densified intensity images are fed into the same pretrained ViNT policy~\citep{shah2023vint} to predict five-step waypoint trajectories, which are evaluated against reference robot trajectories. Across all conditions, Cyclops reduces the average/final waypoint errors from 0.601/1.031~m to 0.140/0.246~m, demonstrating that its generated RGB is compatible with an existing vision-based navigation policy rather than merely improving appearance metrics.

\subsection{Point Cloud Colorization}
\label{subsec:pc_color}

Since each pixel in the colorized image corresponds to a known 3D LiDAR point via the projection mapping, predicted RGB values can be directly assigned to the original point cloud without additional registration. As shown in Fig.~\ref{fig:task}(d), our colorized point clouds exhibit significantly richer visual information: object boundaries are sharper, and surface textures become clearly discernible. At inference, the pipeline requires only LiDAR input and uses the fixed projection mapping established during system setup, without relying on live camera images or additional online calibration.
\section{Limitations and Discussion}
\label{sec:limitations}

Despite the promising results, several limitations remain. The mapping from single-channel intensity to three-channel RGB is inherently one-to-many, meaning the model may hallucinate incorrect colors when distinct materials share similar reflectance profiles. This one-to-many ambiguity is a widely recognized challenge faced by most cross-modal generation frameworks. Additionally, highly dynamic scenes with fast-moving objects or significant ego-motion can lead to degraded colorization quality. We provide a comprehensive discussion of failure cases and dynamic object impact in Appendix~\ref{app:failure_cases}. Future work may address these issues by incorporating per-pixel uncertainty quantification, fusing complementary sensor modalities, or exploiting auxiliary LiDAR outputs. We may also explore a unified multimodal \emph{LiDAR foundation model} that natively integrates RGB color sensing with LiDAR range~\citep{liu2023bevfusion,wang2025vggt}, intensity~\citep{dai2022lidar,gao2026super}, and Doppler velocity~\citep{hexsel2022dicp,gu2022learning,zhao2024fmcw,zuo2026dpnet,narasimhan2026aevascenes}. Combining the strengths of passive and active optical sensing, such a LiDAR foundation model would unify geometric, radiometric, frequency- and phase-based sensing abilities to jointly represent all appearance, metric geometry, spectral characteristics, and differential motion-aware information.
\section{Conclusion}
\label{sec:conclusion}

We present \textit{Cyclops}, a perception-oriented system that produces geometry-conditioned RGB video from sparse NRS-LiDAR intensity alone, enabling RGB-pretrained perception models to operate under LiDAR-only sensing. Our two-stage pipeline integrates intensity image densification with LBM, temporal conditioning, and differentiable reward-guided trajectory optimization to produce temporally stable RGB streams suitable for near-real-time robotic deployment. Experiments across semantic segmentation, lane detection, visual navigation, and point-cloud colorization demonstrate that these RGB streams support standard vision models, outperforming LiDAR baselines in semantic expressiveness and maintaining robust perception when conventional cameras degrade under adverse lighting, thereby demonstrating the system’s practical value in challenging illumination conditions.



\clearpage
\acknowledgments{
This work was supported by the Science and Technology Development Fund
of the Macao Special Administrative Region (FDCT) under Grant Nos.
0067/2023/AFJ and 0117/2024/RIB2.
}

\bibliography{main}  

@inproceedings{shan2021robust,
  title={Robust place recognition using an imaging LiDAR},
  author={Shan et al, Tixiao},
  booktitle={ICRA},
  pages={5469--5475},
  year={2021}
}

@inproceedings{pfreundschuh2024coin,
  title={Coin-LIO: Complementary intensity-augmented LiDAR inertial odometry},
  author={Pfreundschuh et al, Patrick},
  booktitle={ICRA},
  pages={1730--1737},
  year={2024}
}

@misc{tusimple-lane-detection,
  title={TuSimple Lane Detection Challenge},
  author = {{TuSimple}},
  howpublished={\url{https://github.com/TuSimple/tusimple-benchmark}},
  year={2017}
}

@inproceedings{gao2024active,
  title={Active loop closure for OSM-guided robotic mapping in large-scale urban environments},
  author={Gao et al, Wei},
  booktitle={IROS},
  pages={12302--12309},
  year={2024}
}

@article{zhang2023ri,
  title={RI-LIO: Reflectivity image assisted tightly-coupled LiDAR-inertial odometry},
  author={Zhang et al, Yanfeng},
  journal={IEEE RA-L},
  volume={8},
  number={3},
  pages={1802--1809},
  year={2023}
}

@inproceedings{wang2020lidar,
  title={Lidar iris for loop-closure detection},
  author={Wang et al, Ying},
  booktitle={IROS},
  pages={5769--5775},
  year={2020}
}

@article{viswanath2024reflectivity,
  title={Reflectivity Is All You Need!: Advancing LiDAR Segmentation},
  author={Viswanath et al, Kasi},
  journal={arXiv preprint},
  year={2024}
}

@inproceedings{di2021visual,
  title={Visual place recognition using LiDAR intensity},
  author={Di Giammarino et al, Luca},
  booktitle={IROS},
  pages={4382--4389},
  year={2021}
}

@article{dai2022lidar,
  title={LiDAR intensity completion: Fully exploiting sensor messages},
  author={Dai et al, Weichen},
  journal={Sensors},
  volume={22},
  number={19},
  pages={7533},
  year={2022}
}

@inproceedings{isola2017image,
  title={Image-to-image translation with conditional adversarial networks},
  author={Isola, Phillip and Zhu, Jun-Yan and Zhou, Tinghui and Efros, Alexei A},
  booktitle={Proceedings of the IEEE conference on computer vision and pattern recognition},
  pages={1125--1134},
  year={2017}
}

@inproceedings{zhu2017unpaired,
  title={Unpaired image-to-image translation using cycle-consistent adversarial networks},
  author={Zhu, Jun-Yan and Park, Taesung and Isola, Phillip and Efros, Alexei A},
  booktitle={Proceedings of the IEEE international conference on computer vision},
  pages={2223--2232},
  year={2017}
}

@inproceedings{park2019semantic,
  title={Semantic image synthesis with spatially-adaptive normalization},
  author={Park, Taesung and Liu, Ming-Yu and Wang, Ting-Chun and Zhu, Jun-Yan},
  booktitle={Proceedings of the IEEE/CVF conference on computer vision and pattern recognition},
  pages={2337--2346},
  year={2019}
}

@inproceedings{wang2018high,
  title={High-resolution image synthesis and semantic manipulation with conditional gans},
  author={Wang, Ting-Chun and Liu, Ming-Yu and Zhu, Jun-Yan and Tao, Andrew and Kautz, Jan and Catanzaro, Bryan},
  booktitle={Proceedings of the IEEE conference on computer vision and pattern recognition},
  pages={8798--8807},
  year={2018}
}

@inproceedings{ho2020denoising,
  title={Denoising diffusion probabilistic models},
  author={Ho, Jonathan and Jain, Ajay and Abbeel, Pieter},
  booktitle={Advances in Neural Information Processing Systems},
  volume={33},
  pages={6840--6851},
  year={2020}
}

@inproceedings{li2023bbdm,
  title={Bbdm: Image-to-image translation with brownian bridge diffusion models},
  author={Li, Bo and Xue, Kaitao and Liu, Bin and Lai, Yu-Kun},
  booktitle={Proceedings of the IEEE/CVF conference on computer vision and pattern Recognition},
  pages={1952--1961},
  year={2023}
}

@inproceedings{sohl2015deep,
  title={Deep unsupervised learning using nonequilibrium thermodynamics},
  author={Sohl-Dickstein, Jascha and Weiss, Eric and Maheswaranathan, Niru and Ganguli, Surya},
  booktitle={International Conference on Machine Learning},
  pages={2256--2265},
  year={2015}
}

@inproceedings{rombach2022high,
  title={High-resolution image synthesis with latent diffusion models},
  author={Rombach, Robin and Blattmann, Andreas and Lorenz, Dominik and Esser, Patrick and Ommer, Bj{\"o}rn},
  booktitle={Proceedings of the IEEE/CVF conference on computer vision and pattern recognition},
  pages={10684--10695},
  year={2022}
}

@inproceedings{zhang2023adding,
  title={Adding conditional control to text-to-image diffusion models},
  author={Zhang, Lvmin and Rao, Anyi and Agrawala, Maneesh},
  booktitle={Proceedings of the IEEE/CVF International Conference on Computer Vision},
  pages={3836--3847},
  year={2023}
}

@article{mou2024t2i,
  title={T2i-adapter: Learning adapters to dig out more controllable ability for text-to-image diffusion models},
  author={Mou, Chong and Wang, Xintao and Xie, Liangbin and Wu, Yanze and Zhang, Jian and Shan, Ying and Li, Yu},
  journal={AAAI Conference on Artificial Intelligence},
  year={2024}
}

@inproceedings{song2020denoising,
  title={Denoising diffusion implicit models},
  author={Song, Jiaming and Meng, Chenlin and Ermon, Stefano},
  booktitle={International Conference on Learning Representations},
  year={2020}
}

@inproceedings{song2023consistency,
  title={Consistency models},
  author={Song, Yang and Dhariwal, Prafulla and Chen, Mark and Sutskever, Ilya},
  booktitle={International Conference on Machine Learning},
  pages={32211--32252},
  year={2023}
}

@article{luo2023latent,
  title={Latent consistency models: Synthesizing high-resolution images with few-step inference},
  author={Luo, Simian and Tan, Yiqin and Patil, Longbo and Gu, Daniel and von Platen, Patrick and Passos, Apolin{\'a}rio and Huang, Leo and Li, Jian and Zhao, Hang},
  journal={arXiv preprint arXiv:2310.04378},
  year={2023}
}

@article{liu2025veila,
  title={Veila: Panoramic LiDAR generation from a monocular RGB image},
  author={Liu, Youquan and Kong, Lingdong and Yang, Weidong and Liang, Ao and Gao, Jianxiong and Wu, Yang and Xu, Xiang and Li, Xin and Li, Linfeng and Chen, Runnan and others},
  journal={arXiv preprint arXiv:2508.03690},
  year={2025}
}

@article{zhang2025dcltv,
  title={DCLTV: An Improved Dual-Condition Diffusion Model for Laser-Visible Image Translation},
  author={Zhang, Xiaoyu and Zhang, Laixian and Guo, Huichao and Zheng, Haijing and Sun, Houpeng and Li, Yingchun and Li, Rong and Luan, Chenglong and Tong, Xiaoyun},
  journal={Sensors},
  volume={25},
  number={3},
  pages={697},
  year={2025},
  publisher={MDPI}
}

@inproceedings{cortinhal2021semantics,
  title={Semantics-aware multi-modal domain translation: From lidar point clouds to panoramic color images},
  author={Cortinhal, Tiago and Kurnaz, Fatih and Aksoy, Eren Erdal},
  booktitle={Proceedings of the IEEE/CVF International Conference on computer vision},
  pages={3032--3048},
  year={2021}
}

@article{cortinhal2024depth,
  title={Depth-and semantics-aware multi-modal domain translation: Generating 3D panoramic color images from LiDAR point clouds},
  author={Cortinhal, Tiago and Aksoy, Eren Erdal},
  journal={Robotics and Autonomous Systems},
  volume={171},
  pages={104583},
  year={2024},
  publisher={Elsevier}
}

@article{xu2025ligencam,
  title={LiGenCam: Reconstruction of Color Camera Images from Multimodal LiDAR Data for Autonomous Driving},
  author={Xu, Minghao and Gu, Yanlei and Goncharenko, Igor and Kamijo, Shunsuke},
  journal={Sensors},
  volume={25},
  number={14},
  pages={4295},
  year={2025},
  publisher={MDPI}
}

@inproceedings{xu2022point,
  title={Point-nerf: Point-based neural radiance fields},
  author={Xu, Qiangeng and Xu, Zexiang and Philip, Julien and Bi, Sai and Shu, Zhixin and Sunkavalli, Kalyan and Neumann, Ulrich},
  booktitle={Proceedings of the IEEE/CVF conference on computer vision and pattern recognition},
  pages={5438--5448},
  year={2022}
}

@inproceedings{huang2023neural,
  title={Neural lidar fields for novel view synthesis},
  author={Huang, Shengyu and Gojcic, Zan and Wang, Zian and Williams, Francis and Kasten, Yoni and Fidler, Sanja and Schindler, Konrad and Litany, Or},
  booktitle={Proceedings of the IEEE/CVF International Conference on Computer Vision},
  pages={18236--18246},
  year={2023}
}

@article{gao2023magicdrive,
  title={Magicdrive: Street view generation with diverse 3d geometry control},
  author={Gao, Ruiyuan and Chen, Kai and Xie, Enze and Hong, Lanqing and Li, Zhenguo and Yeung, Dit-Yan and Xu, Qiang},
  journal={arXiv preprint arXiv:2310.02601},
  year={2023}
}

@inproceedings{wen2024panacea,
  title={Panacea: Panoramic and controllable video generation for autonomous driving},
  author={Wen, Yuqing and Zhao, Yucheng and Liu, Yingfei and Jia, Fan and Wang, Yanhui and Luo, Chong and Zhang, Chi and Wang, Tiancai and Sun, Xiaoyan and Zhang, Xiangyu},
  booktitle={Proceedings of the IEEE/CVF Conference on Computer Vision and Pattern Recognition},
  pages={6902--6912},
  year={2024}
}

@article{gao2026super,
  title={Super LiDAR Intensity for Robotic Perception},
  author={Gao, Wei and Zhang, Jie and Zhao, Mingle and Zhang, Zhiyuan and Kong, Shu and Ghaffari, Maani and Song, Dezhen and Xu, Chengzhong and Kong, Hui},
  journal={IEEE Robotics and Automation Letters},
  year={2026},
  publisher={IEEE}
}

@inproceedings{lee2022patchwork++,
  title={Patchwork++: Fast and robust ground segmentation solving partial under-segmentation using 3D point cloud},
  author={Lee, Seungjae and Lim, Hyungtae and Myung, Hyun},
  booktitle={2022 IEEE/RSJ International Conference on Intelligent Robots and Systems (IROS)},
  pages={13276--13283},
  year={2022},
  organization={IEEE}
}

@article{steinke2023groundgrid,
  title={Groundgrid: Lidar point cloud ground segmentation and terrain estimation},
  author={Steinke, Nicolai and Goehring, Daniel and Rojas, Ra{\'u}l},
  journal={IEEE Robotics and Automation Letters},
  volume={9},
  number={1},
  pages={420--426},
  year={2023},
  publisher={IEEE}
}

@inproceedings{zhang2018unreasonable,
  title={The unreasonable effectiveness of deep features as a perceptual metric},
  author={Zhang, Richard and Isola, Phillip and Efros, Alexei A and Shechtman, Eli and Wang, Oliver},
  booktitle={Proceedings of the IEEE conference on computer vision and pattern recognition},
  pages={586--595},
  year={2018}
}

@article{heusel2017gans,
  title={Gans trained by a two time-scale update rule converge to a local nash equilibrium},
  author={Heusel, Martin and Ramsauer, Hubert and Unterthiner, Thomas and Nessler, Bernhard and Hochreiter, Sepp},
  journal={Advances in neural information processing systems},
  volume={30},
  year={2017}
}

@inproceedings{chadebec2025lbm,
  title={Lbm: Latent bridge matching for fast image-to-image translation},
  author={Chadebec, Cl{\'e}ment and Tasar, Onur and Sreetharan, Sanjeev and Aubin, Benjamin},
  booktitle={Proceedings of the IEEE/CVF International Conference on Computer Vision},
  pages={29086--29098},
  year={2025}
}

@inproceedings{zhang2025towards,
  title={Towards robust sensor-fusion ground SLAM: A comprehensive benchmark and a resilient framework},
  author={Zhang, Deteng and Zhang, Junjie and Sun, Yan and Li, Tao and Yin, Hao and Xie, Hongzhao and Yin, Jie},
  booktitle={2025 IEEE/RSJ International Conference on Intelligent Robots and Systems (IROS)},
  pages={8894--8901},
  year={2025},
  organization={IEEE}
}

@inproceedings{podell2024sdxl,
  title={Sdxl: Improving latent diffusion models for high-resolution image synthesis},
  author={Podell, Dustin and English, Zion and Lacey, Kyle and Blattmann, Andreas and Dockhorn, Tim and M{\"u}ller, Jonas and Penna, Joe and Rombach, Robin},
  booktitle={International Conference on Learning Representations},
  volume={2024},
  pages={1862--1874},
  year={2024}
}

@inproceedings{yang2025cogvideox,
  title={Cogvideox: Text-to-video diffusion models with an expert transformer},
  author={Yang, Zhuoyi and Teng, Jiayan and Zheng, Wendi and Ding, Ming and Huang, Shiyu and Xu, Jiazheng and Yang, Yuanming and Hong, Wenyi and Zhang, Xiaohan and Feng, Guanyu and others},
  booktitle={International Conference on Learning Representations},
  volume={2025},
  pages={83048--83077},
  year={2025}
}

@article{wang2018video,
  title={Video-to-video synthesis},
  author={Wang, Ting-Chun and Liu, Ming-Yu and Zhu, Jun-Yan and Liu, Guilin and Tao, Andrew and Kautz, Jan and Catanzaro, Bryan},
  journal={arXiv preprint arXiv:1808.06601},
  year={2018}
}

@inproceedings{blattmann2023align,
  title={Align your latents: High-resolution video synthesis with latent diffusion models},
  author={Blattmann, Andreas and Rombach, Robin and Ling, Huan and Dockhorn, Tim and Kim, Seung Wook and Fidler, Sanja and Kreis, Karsten},
  booktitle={Proceedings of the IEEE/CVF conference on computer vision and pattern recognition},
  pages={22563--22575},
  year={2023}
}

@article{ho2022video,
  title={Video diffusion models},
  author={Ho, Jonathan and Salimans, Tim and Gritsenko, Alexey and Chan, William and Norouzi, Mohammad and Fleet, David J},
  journal={Advances in neural information processing systems},
  volume={35},
  pages={8633--8646},
  year={2022}
}

@inproceedings{zhang2016colorful,
  title={Colorful image colorization},
  author={Zhang, Richard and Isola, Phillip and Efros, Alexei A},
  booktitle={European conference on computer vision},
  pages={649--666},
  year={2016},
  organization={Springer}
}

@inproceedings{shen2023style,
  title={Style transfer meets super-resolution: Advancing unpaired infrared-to-visible image translation with detail enhancement},
  author={Shen, Yirui and Kang, Jingxuan and Li, Shuang and Yu, Zhenjie and Wang, Shuigen},
  booktitle={Proceedings of the 31st ACM International Conference on Multimedia},
  pages={4340--4348},
  year={2023}
}

@inproceedings{berg2018generating,
  title={Generating visible spectrum images from thermal infrared},
  author={Berg, Amanda and Ahlberg, Jorgen and Felsberg, Michael},
  booktitle={Proceedings of the IEEE Conference on Computer Vision and Pattern Recognition Workshops},
  pages={1143--1152},
  year={2018}
}

@article{du2023real,
  title={Real-time simultaneous localization and mapping with LiDAR intensity},
  author={Du, Wenqiang and Beltrame, Giovanni},
  journal={arXiv preprint arXiv:2301.09257},
  year={2023}
}

@inproceedings{ha2025enhancing,
  title={Enhancing LiDAR point cloud sampling via colorization and super-resolution of LiDAR imagery},
  author={Ha, Sier and Du, Honghao and Yu, Xianjia and Westerlund, Tomi},
  booktitle={2025 European Conference on Mobile Robots (ECMR)},
  pages={1--7},
  year={2025},
  organization={IEEE}
}

@article{liu20232,
  title = {{I$^{2}$SB}: Image-to-image {Schr{\"o}dinger} bridge},
  author={Liu, Guan-Horng and Vahdat, Arash and Huang, De-An and Theodorou, Evangelos A and Nie, Weili and Anandkumar, Anima},
  journal={arXiv preprint arXiv:2302.05872},
  year={2023}
}

@article{graikos2022diffusion,
  title={Diffusion models as plug-and-play priors},
  author={Graikos, Alexandros and Malkin, Nikolay and Jojic, Nebojsa and Samaras, Dimitris},
  journal={Advances in Neural Information Processing Systems},
  volume={35},
  pages={14715--14728},
  year={2022}
}

@article{liu2022flow,
  title={Flow straight and fast: Learning to generate and transfer data with rectified flow},
  author={Liu, Xingchao and Gong, Chengyue and Liu, Qiang},
  journal={arXiv preprint arXiv:2209.03003},
  year={2022}
}

@article{wu2024moving,
  title={Moving event detection from LiDAR point streams},
  author={Wu, Huajie and Li, Yihang and Xu, Wei and Kong, Fanze and Zhang, Fu},
  journal={nature communications},
  volume={15},
  number={1},
  pages={345},
  year={2024},
  publisher={Nature Publishing Group UK London}
}

@inproceedings{ravi2025sam,
  title={Sam 2: Segment anything in images and videos},
  author={Ravi, Nikhila and Gabeur, Valentin and Hu, Yuan-Ting and Hu, Ronghang and Ryali, Chaitanya and Ma, Tengyu and Khedr, Haitham and R{\"a}dle, Roman and Rolland, Chloe and Gustafson, Laura and others},
  booktitle={International Conference on Learning Representations},
  volume={2025},
  pages={28085--28128},
  year={2025}
}

@inproceedings{tabelini2021keep,
  title={Keep your eyes on the lane: Real-time attention-guided lane detection},
  author={Tabelini, Lucas and Berriel, Rodrigo and Paixao, Thiago M and Badue, Claudine and De Souza, Alberto F and Oliveira-Santos, Thiago},
  booktitle={Proceedings of the IEEE/CVF conference on computer vision and pattern recognition},
  pages={294--302},
  year={2021}
}

@inproceedings{he2016deep,
  title={Deep residual learning for image recognition},
  author={He, Kaiming and Zhang, Xiangyu and Ren, Shaoqing and Sun, Jian},
  booktitle={Proceedings of the IEEE conference on computer vision and pattern recognition},
  pages={770--778},
  year={2016}
}

@article{shah2023vint,
  title={ViNT: A foundation model for visual navigation},
  author={Shah, Dhruv and Sridhar, Ajay and Dashora, Nitish and Stachowicz, Kyle and Black, Kevin and Hirose, Noriaki and Levine, Sergey},
  journal={arXiv preprint arXiv:2306.14846},
  year={2023}
}

@inproceedings{xu2025ctrlora,
  title={Ctrlora: An extensible and efficient framework for controllable image generation},
  author={Xu, Yifeng and He, Zhenliang and Shan, Shiguang and Chen, Xilin},
  booktitle={International Conference on Learning Representations},
  volume={2025},
  pages={19518--19540},
  year={2025}
}

@inproceedings{tan2025ominicontrol,
  title={Ominicontrol: Minimal and universal control for diffusion transformer},
  author={Tan, Zhenxiong and Liu, Songhua and Yang, Xingyi and Xue, Qiaochu and Wang, Xinchao},
  booktitle={2025 IEEE/CVF International Conference on Computer Vision (ICCV)},
  pages={14940--14950},
  year={2025},
  organization={IEEE}
}

@inproceedings{clark2024directly,
  title={Directly fine-tuning diffusion models on differentiable rewards},
  author={Clark, Kevin and Vicol, Paul and Swersky, Kevin and Fleet, David},
  booktitle={International Conference on Learning Representations},
  volume={2024},
  pages={4793--4822},
  year={2024}
}

@article{xu2025frnet,
  title={Frnet: Frustum-range networks for scalable lidar segmentation},
  author={Xu, Xiang and Kong, Lingdong and Shuai, Hui and Liu, Qingshan},
  journal={IEEE Transactions on Image Processing},
  volume={34},
  pages={2173--2186},
  year={2025},
  publisher={IEEE}
}

@inproceedings{ando2023rangevit,
  title={Rangevit: Towards vision transformers for 3d semantic segmentation in autonomous driving},
  author={Ando, Angelika and Gidaris, Spyros and Bursuc, Andrei and Puy, Gilles and Boulch, Alexandre and Marlet, Renaud},
  booktitle={2023 IEEE/CVF Conference on Computer Vision and Pattern Recognition (CVPR)},
  pages={5240--5250},
  year={2023},
  organization={IEEE}
}

@article{zhao2024fmcw,
  title={Fmcw-lio: A doppler lidar-inertial odometry},
  author={Zhao, Mingle and Wang, Jiahao and Gao, Tianxiao and Xu, Chengzhong and Kong, Hui},
  journal={IEEE Robotics and Automation Letters},
  volume={9},
  number={6},
  pages={5727--5734},
  year={2024},
  publisher={IEEE}
}

@article{hexsel2022dicp,
  title={DICP: Doppler iterative closest point algorithm},
  author={Hexsel, Bruno and Vhavle, Heethesh and Chen, Yi},
  journal={arXiv preprint arXiv:2201.11944},
  year={2022}
}

@inproceedings{gu2022learning,
  title={Learning moving-object tracking with fmcw lidar},
  author={Gu, Yi and Cheng, Hongzhi and Wang, Kafeng and Dou, Dejing and Xu, Chengzhong and Kong, Hui},
  booktitle={2022 IEEE/RSJ International Conference on Intelligent Robots and Systems (IROS)},
  pages={3747--3753},
  year={2022},
  organization={IEEE}
}

@article{zuo2026dpnet,
  title={DPNet: Doppler LiDAR Motion Planning for Highly-Dynamic Environments},
  author={Zuo, Wei and Ren, Zeyi and Li, Chengyang and Wang, Yikun and Zhao, Mingle and Wang, Shuai and Sui, Wei and Gao, Fei and Wu, Yik-Chung and Xu, Chengzhong},
  journal={IEEE Robotics and Automation Letters},
  year={2026},
  publisher={IEEE}
}

@article{narasimhan2026aevascenes,
  title={AevaScenes: An FMCW LiDAR Dataset and Benchmark for Long-Range Perception},
  author={Narasimhan, Gautham Narayan and Vhavle, Heethesh and Viswanatha, Kumar Bhargav and Reuther, James and Ramanan, Deva},
  journal={arXiv preprint arXiv:\linebreak[0]2609.33230},
  year={2026}
}

@article{gao2025night,
  title={Night-voyager: Consistent and efficient nocturnal vision-aided state estimation in object maps},
  author={Gao, Tianxiao and Zhao, Mingle and Xu, Chengzhong and Kong, Hui},
  journal={IEEE Transactions on Robotics},
  volume={41},
  pages={2105--2126},
  year={2025},
  publisher={IEEE}
}

@article{cadena2016past,
  title={Past, present, and future of simultaneous localization and mapping: Toward the robust-perception age},
  author={Cadena, Cesar and Carlone, Luca and Carrillo, Henry and Latif, Yasir and Scaramuzza, Davide and Neira, Jos{\'e} and Reid, Ian and Leonard, John J},
  journal={IEEE Transactions on robotics},
  volume={32},
  number={6},
  pages={1309--1332},
  year={2016},
  publisher={IEEE}
}

@inproceedings{wang2025vggt,
  title={Vggt: Visual geometry grounded transformer},
  author={Wang, Jianyuan and Chen, Minghao and Karaev, Nikita and Vedaldi, Andrea and Rupprecht, Christian and Novotny, David},
  booktitle={2025 IEEE/CVF Conference on Computer Vision and Pattern Recognition (CVPR)},
  pages={5294--5306},
  year={2025},
  organization={IEEE}
}

@inproceedings{liu2023bevfusion,
  title={Bevfusion: Multi-task multi-sensor fusion with unified bird's-eye view representation},
  author={Liu, Zhijian and Tang, Haotian and Amini, Alexander and Yang, Xinyu and Mao, Huizi and Rus, Daniela L and Han, Song},
  booktitle={2023 IEEE international conference on robotics and automation (ICRA)},
  pages={2774--2781},
  year={2023},
  organization={ieee}
}

\clearpage
\appendix
\appendix

\paragraph{Appendix Overview} 
This supplementary material is organized into three parts. Appendix~\ref{sec:supp_method} provides additional details on the densification architecture, temporal conditioning, reward-guided optimization, and training configurations. Appendix~\ref{app:experiments} presents dataset details, annotation protocols, baseline analysis, full quantitative results, and limitation discussion. Appendix~\ref{sec:supp_exp} includes an ablation on inference steps, qualitative results on a public dataset, and visualizations of illumination-invariant colorization under varying lighting conditions.

\section{Supplementary Material: Method Details}
\label{sec:supp_method}


This section provides methodology details in Section~\ref{sec:methodology}, with extended discussions on Stage~I generalization and architecture (Appendix~\ref{app:stage1}), temporal conditioning formulations (Appendix~\ref{app:temporal_attn}), reward-guided optimization design (Appendix~\ref{app:reward_extended}), and training details (Appendix~\ref{app:training_details}).

\subsection{Stage I: Intensity Image Densification}
\label{app:stage1}

\subsubsection{Generalization to Other LiDAR Types}
\label{app:generalization}

Our pipeline is developed and trained primarily on data from a Livox MID-360, a representative non-repetitive scanning (NRS) LiDAR. We select this low-cost sensor because its non-repeating scan trajectory allows dense, high-resolution supervision to be constructed through stationary accumulation, without requiring expensive terrestrial laser scanners or labor-intensive manual annotation. Nevertheless, the resulting Stage-I densifier is initially specialized to the field of view, sampling pattern, and intensity distribution of the Livox MID-360.

\begin{figure}
    \centering
    \includegraphics[width=\linewidth]{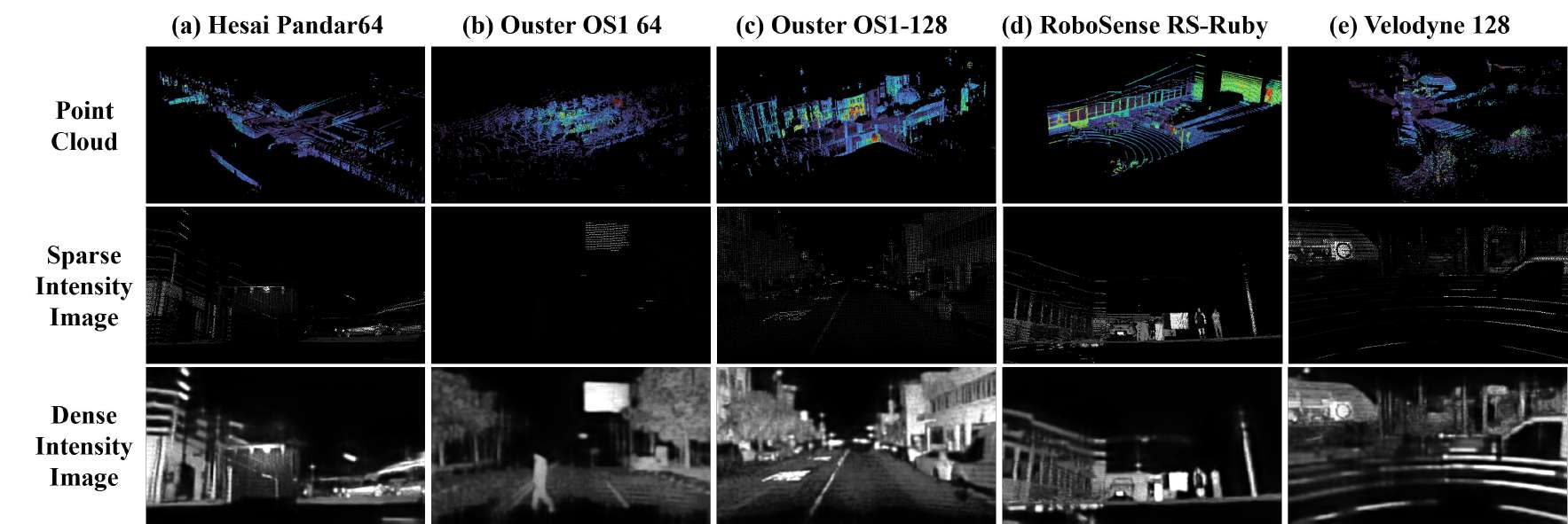}
    \caption{Qualitative results of training-free cross-LiDAR adaptation for Stage~I. From top to bottom: input sparse point clouds, sparse intensity projections after alignment with the Livox MID-360 input domain, and dense intensity images produced by the frozen densification network. The same Stage-I model is applied to multiple LiDARs with different scan patterns and resolutions without sensor-specific retraining or finetuning.}
    \label{fig:cross_lidar}
\end{figure}

\paragraph{Densification Stage.}
To extend Stage~I to other LiDAR types, we introduce a training-free cross-LiDAR adaptation that maps scans from different sensors to the original Livox input domain. Specifically, the projected field of view, sampling sparsity, and intensity statistics are aligned before being processed by the frozen densification network. This adaptation is implemented entirely as input preprocessing and requires neither network-parameter updates nor paired RGB supervision. As shown in Fig.~\ref{fig:cross_lidar}, the adapted inputs from different scan patterns can be processed by the original Stage-I network to produce dense intensity images without sensor-specific retraining.

More generally, sensor-specific dense supervision could also be constructed when stronger adaptation is required. For spinning LiDARs, repeated scans from a stationary platform produce nearly identical scan lines because of their fixed scanning patterns, limiting coverage diversity. Possible alternatives include: (i) exploiting ego-motion during slow driving to obtain parallax-induced coverage, (ii) fusing multi-beam returns from slightly different sensor poses through SLAM-based registration, or (iii) constructing dense ground truth using high-resolution terrestrial laser scanning co-registered with the LiDAR frame. These strategies provide complementary ways to adapt the training data to different scan geometries while retaining the same densification architecture.

\paragraph{Colorization Stage.}
Stage~II operates on dense 2D intensity images rather than raw point clouds or sensor-specific scan patterns, allowing colorization to be performed in a common image domain. This design separates the colorization network architecture from the sensor-dependent organization. Once Stage~I produces a sufficiently complete intensity image in the expected representation, the same colorization network can be used without architectural modification.

\subsubsection{Densification Network Architecture}
\label{app:stage1_arch}

The densification network $\mathcal{F}_{\mathrm{dense}}$ from~\citep{gao2026super} adopts a U-shaped encoder--decoder architecture augmented with two task-specific modules (as illustrated in Section III.C of~\citep{gao2026super}):

\paragraph{Adaptive Fusion Module (AFM).}
The AFM employs parallel dilated convolutions at varying dilation rates together with deformable convolutions to expand the receptive field and adaptively aggregate multi-scale context. This enables the network to recover structural details in large void regions while preserving high-frequency edges.

\paragraph{Dynamic Compensation Module (DCM).}
The DCM takes the decoder prediction together with geometric cues such as range and incidence angle, and produces a physically calibrated dense intensity image. This compensates for intensity variations caused by distance-dependent attenuation and surface orientation effects.

The network is supervised by an MSE reconstruction loss and trained on the \textit{Super LiDAR Intensity} dataset~\citep{gao2026super}, which provides paired sparse--dense intensity images collected by a stationary Livox MID-360 across 20 urban scenes.

\subsection{Temporal Conditioning: Detailed Formulations}
\label{app:temporal_attn}

\paragraph{Cross Attention (Source Conditioning).}
At each spatial resolution level of the U-Net, the current intermediate feature $h$ serves as the query, while the source latent $z^{\mathrm{src}}_t$ provides the key and value:
\begin{equation}
    h' = h + \mathrm{CrossAttn}\!\left(
    Q = W_q h,\;
    K = W_k z^{\mathrm{src}}_t,\;
    V = W_v z^{\mathrm{src}}_t
    \right),
    \label{eq:cross_attn_supp}
\end{equation}
where $W_q$, $W_k$, and $W_v$ are learnable projection matrices. Before attention, the feature maps and latent maps are projected into token sequences with matched channel dimensions. This mechanism enables the velocity field to attend to the fine-grained spatial structure of the source intensity at every resolution scale, ensuring that the predicted velocity respects the underlying scene geometry.

\paragraph{Temporal Attention (History Conditioning).}
At each resolution level, the cross-attention-enhanced feature $h'$ serves as the query, while $z_{\mathrm{prev}}$ provides the key and value:
\begin{equation}
    h'' = h' + \mathrm{TempAttn}\!\left(
    Q = W_q' h',\;
    K = W_k' z_{\mathrm{prev}},\;
    V = W_v' z_{\mathrm{prev}}
    \right),
    \label{eq:temporal_attn_supp}
\end{equation}
where $W_q'$, $W_k'$, and $W_v'$ are learnable projection matrices independent of the cross-attention layers. During teacher-forced training, $z_{\mathrm{prev}}=z^{\mathrm{gt}}_{t-1}$; during model-conditioned training and inference, $z_{\mathrm{prev}}=\hat{z}^{\mathrm{tgt}}_{t-1}$. For the first frame, the temporal attention branch is replaced by the learnable null token $z_{\varnothing}$, providing neutral temporal context without imposing artificial previous-frame appearance.

At test time, no ground-truth RGB frame is used, and the temporal memory is updated autoregressively using the previously generated latent. Importantly, the temporal branch acts as an appearance-memory module rather than a pixel-wise fusion module. Since temporal information is injected through attention instead of direct addition or concatenation, the model does not assume exact spatial alignment between consecutive frames. The source cross-attention remains responsible for anchoring the generation to the current-frame LiDAR intensity geometry, while temporal attention promotes color persistence and reduces flickering by retrieving relevant appearance cues from the previous target latent.
\subsection{Reward-Guided Trajectory Optimization: Extended Discussion}
\label{app:reward_extended}

\paragraph{Sequential Decision Process Interpretation.}
To motivate our reward design, we interpret the intra-frame ODE generation as a sequential decision process:
\begin{itemize}[nosep,leftmargin=*]
    \item \textbf{State}: $s_{t,m}=(z_{t,m},\, z^{\mathrm{src}}_t,\, z_{\mathrm{prev}})$, containing the current intermediate latent, the source intensity latent, and the previous-frame target latent at step $m$.
    \item \textbf{Action}: The velocity prediction $a_{t,m}=v_\theta(z_{t,m},m/M,c_t)$, which determines the latent transition direction and magnitude.
    \item \textbf{Transition}: The deterministic Euler update $z_{t,m+1}=z_{t,m}+a_{t,m}\Delta\tau$.
    \item \textbf{Reward}: A terminal reward $\mathcal{R}(t)$ evaluated at the final step $m=M$, assessing the quality of the generated terminal latent $\hat{z}^{\mathrm{tgt}}_t$.
\end{itemize}

Unlike conventional reinforcement learning, which typically relies on stochastic policy gradient estimators to handle non-differentiable dynamics or exploration requirements, our pipeline is fully differentiable. Therefore, the terminal reward can be optimized by direct backpropagation through the deterministic ODE solver, making our approach an instance of \emph{differentiable reward-guided trajectory optimization} rather than a stochastic RL algorithm.

As illustrated in Fig.~\ref{fig:framework}(b), the BPTT signal flows from the terminal reward function back through the entire ODE solver chain (i.e., through all intermediate latent states $z_{t,0}\rightarrow z_{t,1}\rightarrow\cdots\rightarrow z_{t,M}$), allowing the velocity policy to account for the long-term consequences of its intermediate predictions and to correct trajectory deviations that would otherwise accumulate over multiple integration steps. Importantly, $z_{\mathrm{prev}}$ is treated as a fixed conditioning input during backpropagation (i.e., its gradient is detached), so that the BPTT remains confined to the current frame's $M$ integration steps and does not propagate into the previous frame's generation chain, thereby avoiding prohibitive memory cost and training instability.

\paragraph{Temporal Reward Design Discussion.}
A natural alternative to our temporal reward (Eq.~\ref{eq:reward_temp} in the main paper) would be to directly minimize $\|\hat{z}^{\mathrm{tgt}}_t - \hat{z}^{\mathrm{tgt}}_{t-1}\|$, which penalizes all inter-frame changes indiscriminately. However, this would suppress legitimate motion and scene changes, producing overly static outputs.

Our formulation instead matches the generated temporal displacement $(\hat{z}^{\mathrm{tgt}}_t - z_{\mathrm{prev}})$ to the ground-truth temporal displacement $(z^{\mathrm{gt}}_t - z^{\mathrm{gt}}_{t-1})$. This preserves real scene dynamics while penalizing abnormal flickering or appearance drift that deviates from the true temporal evolution.

The $z_{\mathrm{prev}}$ term follows the same scheduled-sampling protocol as the temporal attention input: under teacher forcing, $z_{\mathrm{prev}}=z^{\mathrm{gt}}_{t-1}$; under model-conditioned training, $z_{\mathrm{prev}}=\hat{z}^{\mathrm{tgt}}_{t-1}$. For the first frame of a sequence, the temporal reward is omitted since no valid previous-frame displacement is available.

We note that under teacher forcing ($z_{\mathrm{prev}}=z^{\mathrm{gt}}_{t-1}$), the temporal reward simplifies to $\mathcal{R}_{\mathrm{temp}}(t) = -\|\hat{z}^{\mathrm{tgt}}_t - z^{\mathrm{gt}}_t\|_1$, which is effectively an L1 fidelity term complementary to the L2-based $\mathcal{R}_{\mathrm{fid}}$. The temporal consistency role of $\mathcal{R}_{\mathrm{temp}}$ is fully realized during model-conditioned training, where $z_{\mathrm{prev}}=\hat{z}^{\mathrm{tgt}}_{t-1}$ introduces the model's own prediction errors into the displacement computation, thereby explicitly encouraging the velocity policy to produce temporally coherent trajectories.

\subsection{Training Details}
\label{app:training_details}

\paragraph{Phase 1: Supervised Pre-training.}
The composite objective is:
\begin{equation}
    \mathcal{L}_{\mathrm{Phase1}}=\mathcal{L}_{\mathrm{LBM}}+\lambda_1\mathcal{L}_{\mathrm{lpips}}+\lambda_2\mathcal{L}_{\mathrm{grad}}+\lambda_3\mathcal{L}_{\mathrm{color}},
\end{equation}
where $\mathcal{L}_{\mathrm{lpips}}$ is the LPIPS perceptual distance~\citep{zhang2018unreasonable} computed on decoded outputs. The spatial gradient loss is defined as:
\begin{equation}
    \mathcal{L}_{\mathrm{grad}} = \|\nabla_x \hat{y}_t - \nabla_x y_t\|_1 + \|\nabla_y \hat{y}_t - \nabla_y y_t\|_1,
\end{equation}
where $\nabla_x$ and $\nabla_y$ denote horizontal and vertical Sobel operators applied to the decoded prediction $\hat{y}_t$ and ground truth $y_t$, respectively. The color statistics loss is:
\begin{equation}
    \mathcal{L}_{\mathrm{color}} = \sum_{c \in \{R,G,B\}} \left( \left|\mu_c(\hat{y}_t) - \mu_c(y_t)\right| + \left|\sigma_c(\hat{y}_t) - \sigma_c(y_t)\right| \right),
\end{equation}
where $\mu_c(\cdot)$ and $\sigma_c(\cdot)$ denote the spatial mean and standard deviation of channel $c$, preventing global color drift.

During this phase, temporal attention uses teacher forcing ($z_{\mathrm{prev}}=z^{\mathrm{gt}}_{t-1}$). For the first frame of a sequence, the temporal attention branch is replaced by the learnable null token $z_{\varnothing}$.

\paragraph{Phase 2: Reward-Driven Fine-tuning.}
After Phase~1 converges, we introduce the terminal reward and switch to scheduled sampling. With probability $p_{\mathrm{tf}}$, the temporal attention module receives $z^{\mathrm{gt}}_{t-1}$; with probability $1-p_{\mathrm{tf}}$, it receives the model's own prediction $\hat{z}^{\mathrm{tgt}}_{t-1}$. The teacher-forcing probability is linearly annealed from $p_{\mathrm{tf}}=1.0$ to $p_{\mathrm{tf}}=0.2$ over the course of Phase~2.

The Phase~2 objective is:
\begin{equation}
    \mathcal{L}_{\mathrm{Phase2}} = \mathcal{L}_{\mathrm{Phase1}} - \lambda_4 \mathcal{R}(t),
\end{equation}
where $\mathcal{R}(t)=\omega_1\mathcal{R}_{\mathrm{fid}}(t)+\omega_2\mathcal{R}_{\mathrm{temp}}(t)$. Since $\mathcal{R}(t)$ is defined as a weighted sum of negative distances, minimizing this loss is equivalent to minimizing the terminal fidelity and temporal consistency penalties. The learning rate of $v_\theta$ is reduced by $10\times$ relative to Phase~1 to preserve the generative prior. Loss weights are set to $\lambda_1=1.0$, $\lambda_2=0.1$, $\lambda_3=0.05$, $\lambda_4=0.8$, and reward coefficients to $\omega_1=1.0$, $\omega_2=0.5$ (tuned on a held-out validation set).

\section{Supplementary Material: Experimental Details}
\label{app:experiments}

This section supplements the experimental evaluation in Section~4 with dataset information, metric definitions, baseline analysis, and full quantitative results.

\subsection{Dataset Details}
\label{app:dataset}

\begin{figure}
    \centering
    \includegraphics[width=0.5\linewidth]{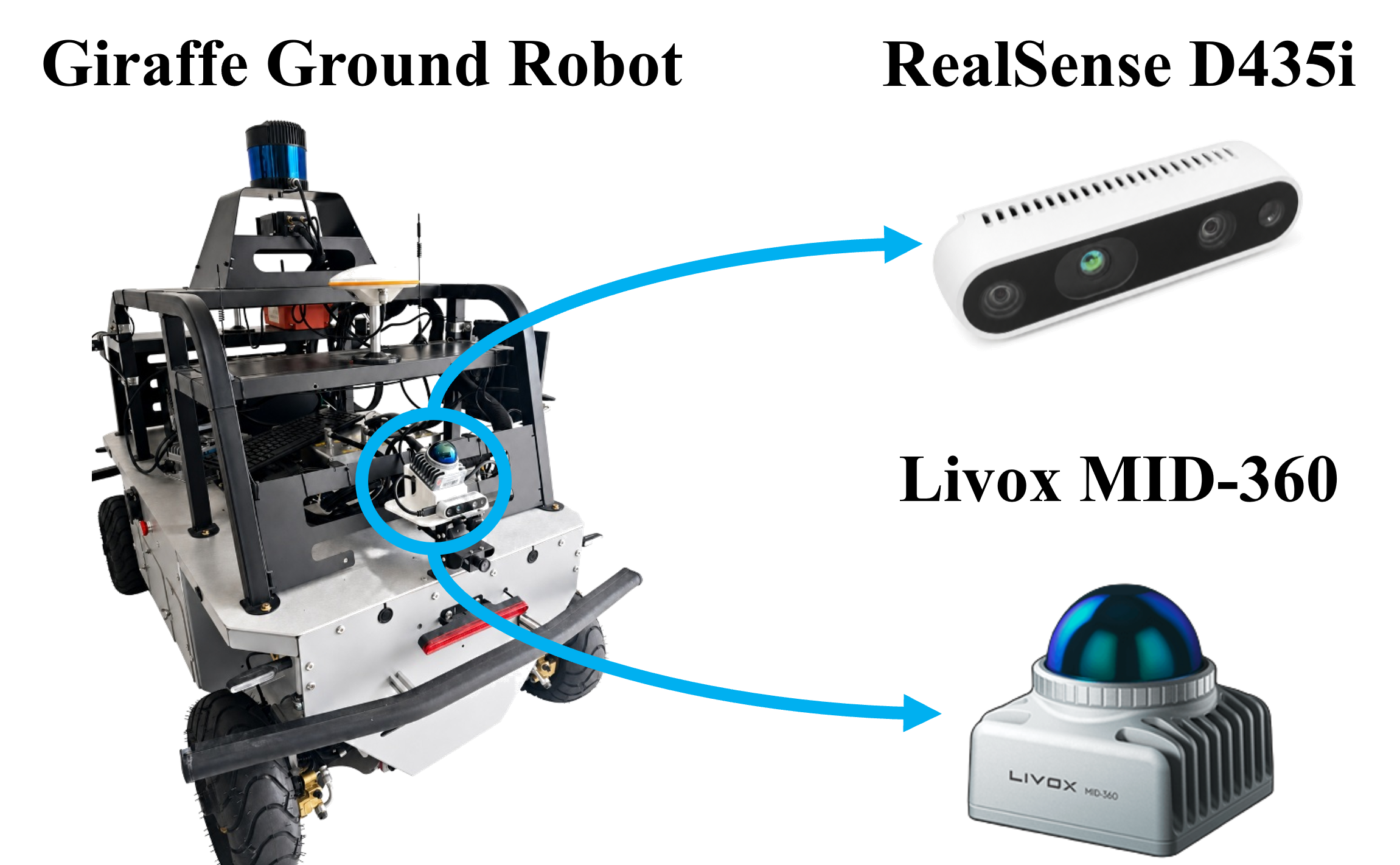}
    \caption{Data collection platform. A ground robot equipped with a Livox MID-360 LiDAR and a RealSense D435i RGB-D camera. The two sensors are rigidly mounted and extrinsically calibrated.}
    \label{fig:robot}
\end{figure}

\paragraph{Data Collection.}
All data are collected using a Livox MID-360 LiDAR and a RealSense D435i camera mounted on a ground robot (as shown in Fig.~\ref{fig:robot}). The extrinsic calibration is obtained via a target-based procedure with sub-pixel reprojection error. The robot traverses environments at speed $\sim$0.5-1.5 m/s, covering diverse urban scenes including gardens, libraries, plazas, etc.

\paragraph{Dataset Component.}
Table~\ref{tab:dataset_split} summarizes the dataset composition. Our dataset comprises a total of 43 sequences (28{,}495 synchronized frame pairs) with three lighting conditions. \textbf{Bright} sequences (34 sequences, including 4 from the public M3DGR dataset~\citep{zhang2025towards} captured with the same Livox MID-360 sensor) are captured under clear daytime conditions and are split into training, validation, and test subsets at an approximate 6:1:1 ratio. \textbf{Night} sequences (9 sequences) are further divided into \textit{Low Light} (4 sequences, captured during dusk/dawn or under heavy overcast with artificial street lighting) and \textit{Near Dark} (5 sequences, captured at night with minimal ambient illumination). The train/val/test split is performed at the sequence level with no geographic overlap: training and test sequences are collected from non-overlapping urban scenes, preventing data leakage through shared scene content.

\begin{table}[t]
\centering
\caption{Dataset composition by lighting condition and split. Training and validation use only Bright-condition data; Low Light and Near Dark data appear exclusively in the test set.}
\label{tab:dataset_split}
\begin{tabular}{lccc}
\toprule
\textbf{Condition} & \textbf{Sequences} & \textbf{Frames} & \textbf{Train / Val / Test} \\
\midrule
Bright         & 34 & 24{,}983 & 18{,}721 / 3{,}308 / 2{,}954 \\
Low Light      & 4  & 1{,}504  & --- / --- / 1{,}504            \\
Near Dark      & 5  & 2{,}008  & --- / --- / 2{,}008            \\
\midrule
\textbf{Total} & 43 & 28{,}495 & 18{,}721 / 3{,}308 / 6{,}466  \\
\bottomrule
\end{tabular}
\end{table}
\paragraph{Segmentation Annotation Protocol.}
Ground-truth semantic masks for the segmentation evaluation (\S\ref{subsec:seg}) are manually annotated on 1000+ test images using the LabelMe tool. Each image is labeled into five semantic categories: \emph{Road}, \emph{Sidewalk}, \emph{Vegetation}, \emph{Building}, and \emph{Lawn}. Annotations are performed at full camera resolution and downsampled to the evaluation resolution ($256\times455$) to preserve label boundaries. For the near-dark and low-light conditions, where camera images are severely degraded, annotators additionally refer to co-registered daytime imagery and LiDAR intensity projections to ensure annotation quality.

\paragraph{Lane Detection Annotation.}
Lane annotations follow the TuSimple format. A maximum of four lanes per image are annotated. We annotate 1{,}800 images total (1{,}300 train / 300 val / 200 test). Since the camera RGB, densified LiDAR intensity image, and predicted RGB are pixel-aligned by construction, a single set of lane annotations suffices for all three modalities, enabling a fair comparison when fine-tuning the lane detection model on each input type independently.

\subsection{Detailed Baseline Analysis}
\label{app:baseline_analysis}

We provide additional details on each baseline method compared in \S\ref{subsec:colorization_eval}. CycleGAN~\citep{zhu2017unpaired} learns unpaired image translation via cycle-consistency loss; the lack of direct supervision leads to color distortion and hallucinated textures, and adversarial training introduces stochastic inter-frame variations. Pix2Pix~\citep{isola2017image} benefits from paired supervision, which improves structural alignment, but the L1 loss combined with a patch-based discriminator produces overly blurry outputs that lose fine structural details. BBDM~\citep{li2023bbdm} formulates translation as a Brownian Bridge diffusion process, achieving the best perceptual quality among baselines; however, stochastic denoising still introduces inter-frame variation. LDM~\citep{rombach2022high} is adapted by concatenating the source latent with the noise input; while outputs are perceptually detailed, independent stochastic sampling per frame yields the worst temporal consistency.

\subsection{Semantic Segmentation: Detailed Setup and Results}
\label{app:seg_setup}

For image-based methods, including SAM\,2 with camera, densified intensity, and Cyclops-rendered RGB inputs, predictions are directly evaluated against the five-class ground-truth masks in image space. For point-cloud methods, per-point predictions are projected onto the camera image plane using the known LiDAR-to-camera calibration. Morphological closing and largest connected component extraction are subsequently applied to obtain dense prediction regions.

\paragraph{Video Point-Prompt Protocol.}
SAM\,2~\citep{ravi2025sam} is evaluated in video-predictor mode rather than independently on each frame. Identical point prompts are provided across all image modalities at five-frame intervals, with 3-8 representative points for each semantic class present in a prompted keyframe. Masks for the intervening unprompted frames are propagated through SAM\,2's temporal memory, and predictions on all frames are evaluated.

\begin{figure}[t]
    \centering
    \includegraphics[width=\linewidth]{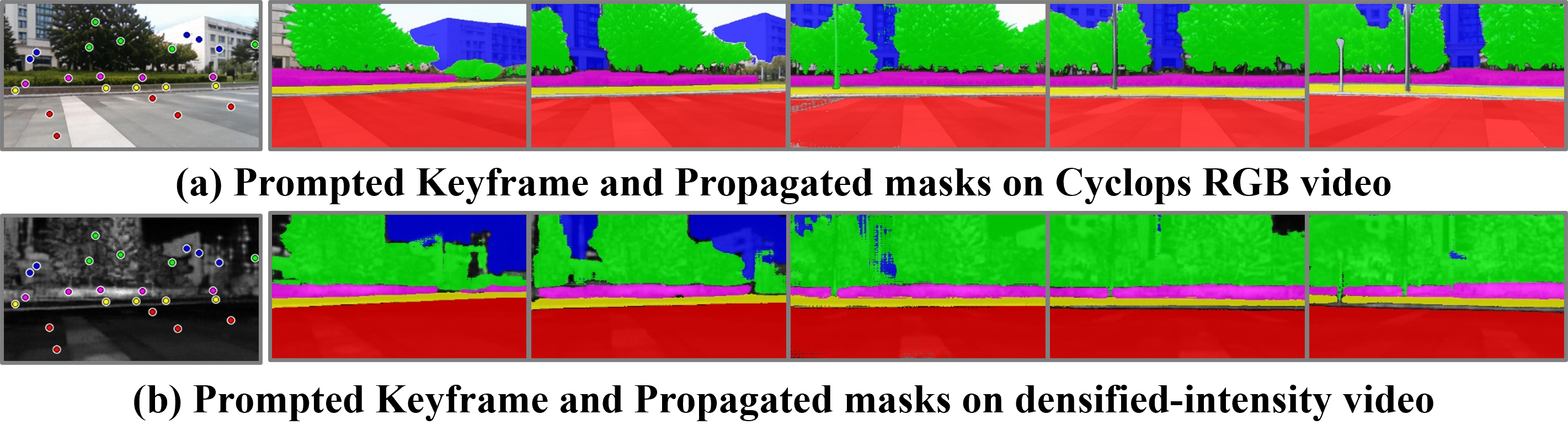}
    \caption{SAM\,2 mask propagation using identical prompts. The first column shows the prompted keyframe, followed by masks propagated to subsequent unprompted frames. Cyclops-rendered RGB produces more complete and temporally stable masks than densified intensity input.}
    \label{fig:sam2_propagation}
\end{figure}

\paragraph{Prompting-Frequency Robustness.}
As illustrated in Fig.~\ref{fig:sam2_propagation}, the semantic cues provided by Cyclops improve mask propagation on unprompted frames. When the prompting interval is increased from five to ten frames, densified intensity obtains 27.6/26.4/26.1 mIoU under Bright/Low-Light/Near-Dark conditions, whereas Cyclops maintains 60.6/58.7/59.3 mIoU, respectively.

\paragraph{Binary LiDAR References.}
Patchwork++~\citep{lee2022patchwork++} and GroundGrid~\citep{steinke2023groundgrid} perform binary ground/non-ground segmentation. For their evaluation, the five annotated classes are grouped into two superclasses: \{Road, Sidewalk, Lawn\} as ground and \{Vegetation, Building\} as non-ground. Pixels outside the valid projected regions are excluded from their metric computation. Their results are therefore included as coarse LiDAR ground-segmentation references rather than equivalent five-class semantic segmentation methods.

\begin{figure}[t]
    \centering
    \includegraphics[width=\linewidth]{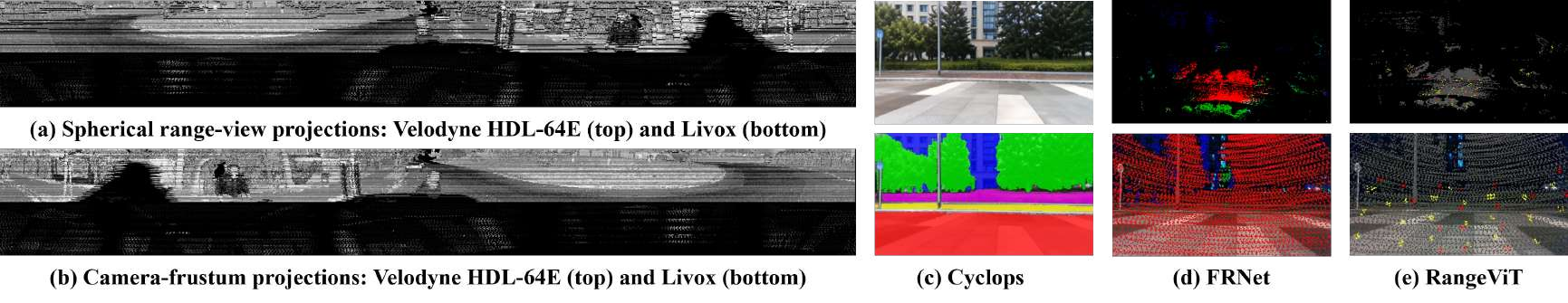}
    \caption{Comparison with projection-based LiDAR segmentation baselines. \textbf{(a--b)} Spherical range-view and camera-frustum projections of regular spinning-LiDAR scans (top) and Livox MID-360 scans (bottom). \textbf{(c--e)} Semantic segmentation results from the Cyclops-based pipeline, FRNet, and RangeViT, respectively.}
    \label{fig:frnet_rangevit}
\end{figure}

\paragraph{Projection-Based LiDAR Baselines.}
We additionally fine-tune FRNet~\citep{xu2025frnet} and RangeViT~\citep{ando2023rangevit} using the same five-scan inputs and five-class annotations. As illustrated in Fig.~\ref{fig:frnet_rangevit}, their range-view and camera-frustum representations assume relatively regular scan geometry and do not naturally accommodate the irregular, time-varying projections of the Livox MID-360. FRNet and RangeViT obtain 16.8 and 5.4 mIoU, respectively, compared with 65.2 mIoU for the Cyclops-based segmentation pipeline under the same evaluation setting.

The main quantitative results under the standard 5-frame prompting protocol are reported in Table~\ref{tab:segmentation}.
\subsection{Traffic Lane Detection: Full Setup and Quantitative Results}
\label{app:lane_details}

\paragraph{Model and Training.}
We adopt LaneATT~\citep{tabelini2021keep} with a ResNet-34 backbone~\citep{he2016deep} as the detection model. The model is first pre-trained on the TuSimple benchmark~\citep{tusimple-lane-detection}, and then fine-tuned on a small set of annotated images from our domain ($\sim$1,300 training images, 300 validation images, and 200 test images). To isolate the effect of input modality, we fine-tune three copies of the same pre-trained model: one on camera images, one on densified intensity images, and one on our colorized LiDAR intensity images, using the same annotations and training schedule.

\paragraph{Evaluation Metrics.}
Detection performance is evaluated using standard TuSimple metrics: Accuracy (the proportion of correctly predicted lane points within a threshold distance), F1 score (harmonic mean of precision and recall at the lane level), False Positive Rate (FPR, proportion of predicted lanes not matching any ground-truth lane), and False Negative Rate (FNR, proportion of ground-truth lanes not detected).

\begin{table}[ht]
    \centering
    \caption{Traffic lane detection results under different lighting conditions.}
    \label{tab:lane_supp}
    \scriptsize
    \setlength{\tabcolsep}{3.5pt}
    \renewcommand{\arraystretch}{1.2}
    \begin{tabular}{@{}ll cccc@{}}
        \toprule
        \textbf{Condition} & \textbf{Input} & \textbf{Acc.(\%)}$\uparrow$ & \textbf{F1(\%)}$\uparrow$ & \textbf{FPR(\%)}$\downarrow$ & \textbf{FNR(\%)}$\downarrow$ \\
        \midrule
        \multirow{3}{*}{Bright}
            & Camera              & \textbf{96.5} & \textbf{96.2} & \textbf{3.1} & \textbf{3.5} \\
            & Densified Intensity Image & 96.1 & 95.2 & 5.2 & 5.5 \\
            & Ours (colorized)    & 95.8 & 95.4 & 3.8 & 4.2 \\
        \midrule
        \multirow{3}{*}{Near Dark}
            & Camera              & 71.2 & 68.5 & 16.3 & 19.8 \\
            & Densified Intensity Image & 93.3 & 92.5 & 6.9 & 6.9 \\
            & Ours (colorized)    & \textbf{95.2} & \textbf{94.7} & \textbf{4.2} & \textbf{4.8} \\
        \midrule
        \multirow{3}{*}{Average}
            & Camera              & 83.9 & 82.4 & 9.7 & 11.7 \\
            & Densified Intensity Image & 94.7 & 93.9 & 6.1 & 6.2 \\
            & Ours (colorized)    & \textbf{95.5} & \textbf{95.1} & \textbf{4.0} & \textbf{4.5} \\
        \bottomrule
    \end{tabular}
\end{table}

\paragraph{Full Results.}
Table~\ref{tab:lane_supp} presents quantitative traffic lane detection results under Bright and Near-Dark conditions with different image types as input.

\subsection{Limitation: Failure Case Analysis}
\label{app:failure_cases}

We identify and analyze the primary failure modes of our pipeline.

\paragraph{Color Ambiguity from Reflectance Degeneracy.}
The most fundamental limitation arises from the one-to-many nature of the intensity-to-RGB mapping: physically distinct materials may exhibit nearly identical LiDAR reflectance yet possess different visual appearances. In such cases, the model tends to regress toward the dataset's modal color for that reflectance range, producing plausible but potentially incorrect colorizations. This ambiguity is a widely recognized challenge shared by cross-modal generation frameworks that synthesize RGB from reduced-channel inputs, including grayscale colorization~\citep{zhang2016colorful} and infrared-to-visible translation~\citep{shen2023style, berg2018generating}.

\paragraph{Far-Field Degradation.}
Distant objects suffer from the compounding effects of reduced intensity signal and sparser point coverage, leaving residual spatial gaps even after densification. The colorization network consequently produces blurred textures and occasional hallucinated structures, particularly for thin structures (e.g., poles, wires) and small objects (e.g., distant pedestrians).

\paragraph{Temporal Consistency Breakdown.}
Despite the temporal attention mechanism and reward-guided optimization, temporal artifacts may still occur under rapid viewpoint changes, rapid ego-motion, or in open scenes with sparse structural context, where the inter-frame correspondence becomes unreliable, and the temporal prior fails to provide effective regularization.

\paragraph{Downstream Impact and Mitigation.}
Color ambiguity has limited impact on tasks that rely primarily on spatial structure, such as segmentation, but may directly affect tasks that require fine-grained color discrimination (e.g., traffic light recognition). Far-field degradation reduces the effective detection range for tasks like lane detection, though the high reflectance of lane markings partially offsets this effect. We believe that incorporating per-pixel uncertainty estimation, fusing complementary modalities (e.g., thermal cameras), leveraging auxiliary LiDAR channels, and extending the temporal context window may help alleviate these limitations.

\section{Supplementary Experiments Results}
\label{sec:supp_exp}
\subsection{Ablation on Inference Steps}
\label{app:nfe_analysis}

\begin{figure}
    \centering
    \includegraphics[width=\linewidth]{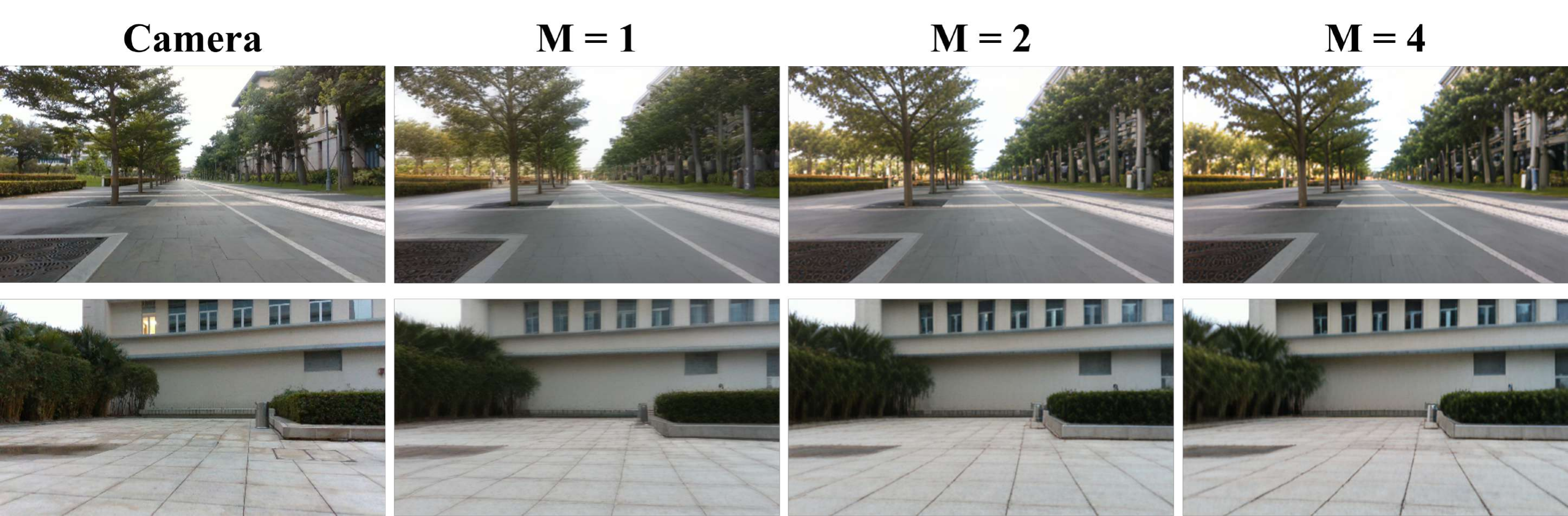}
    \caption{Qualitative comparison under different numbers of function evaluations (NFE). The model is trained exclusively with \(M=4\). From left to right: the ground-truth camera image and outputs with \(M\in\{1,2,4\}\).}
    \label{fig:nfe_comparison}
\end{figure}

Our velocity field is trained with \(M=4\) equally spaced bridge time steps, which serve as implicit distillation targets (cf.\ \S\ref{subsec:lbm}). To examine the model's sensitivity to inference-time step count, we fix the trained model and vary the number of Euler steps at inference among \(M\in\{1,2,4\}\). As shown in Fig.~\ref{fig:nfe_comparison}, \(M=4\) produces the most visually coherent results, while fewer steps lead to progressive degradation in color accuracy and texture sharpness.

\subsection{Performance on Public Dataset Sequence}
\label{app:public_performance}

\begin{figure}
    \centering
    \includegraphics[width=\linewidth]{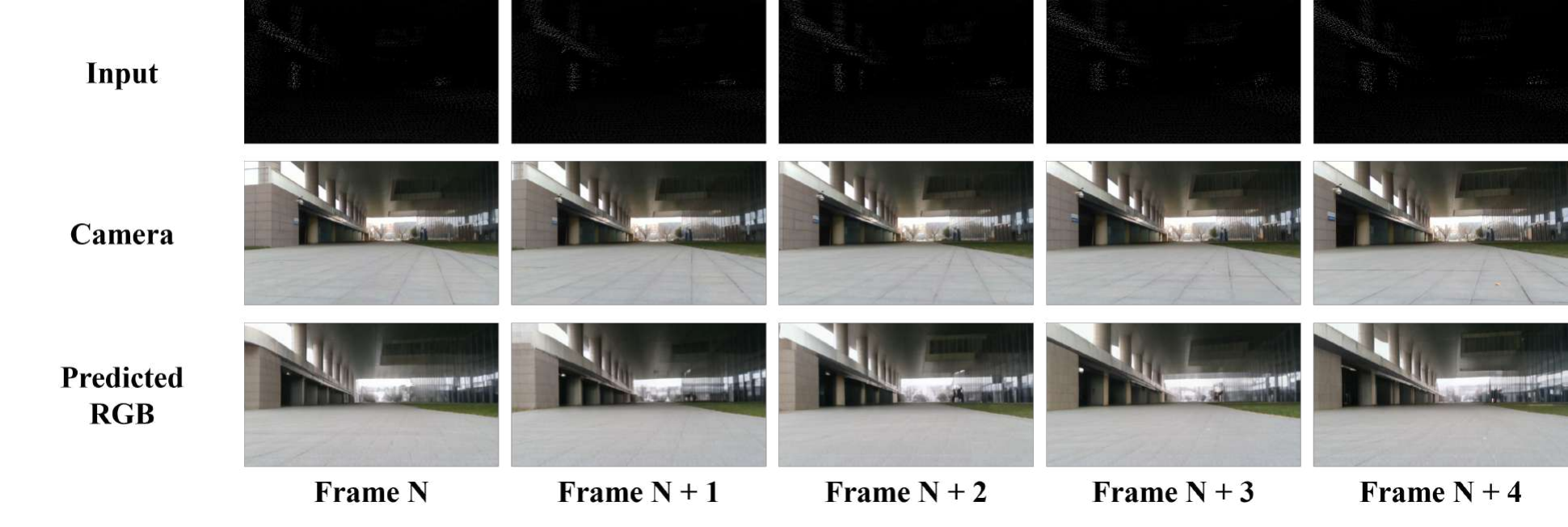}
    \caption{An example of consecutive frames generated by our method on the GNSS-denial02 sequence from the public M3DGR dataset~\citep{zhang2025towards}.}
    \label{fig:publiC_performance}
\end{figure}

To evaluate the generalization ability of our method, we qualitatively test on the GNSS-denial01 and GNSS-denial02 sequences from the public M3DGR dataset~\citep{zhang2025towards}. Both sequences are captured in open outdoor environments where LiDAR returns are notably sparser, especially at long range, due to the lack of nearby structures. As a result, the generated images exhibit some degradation in texture detail compared to our self-collected scenes. Nevertheless, as shown in Fig.~\ref{fig:publiC_performance}, our method still preserves plausible geometric structure and overall scene layout, demonstrating reasonable robustness under challenging sparse-input conditions.

\begin{figure}
    \centering
    \includegraphics[width=\linewidth]{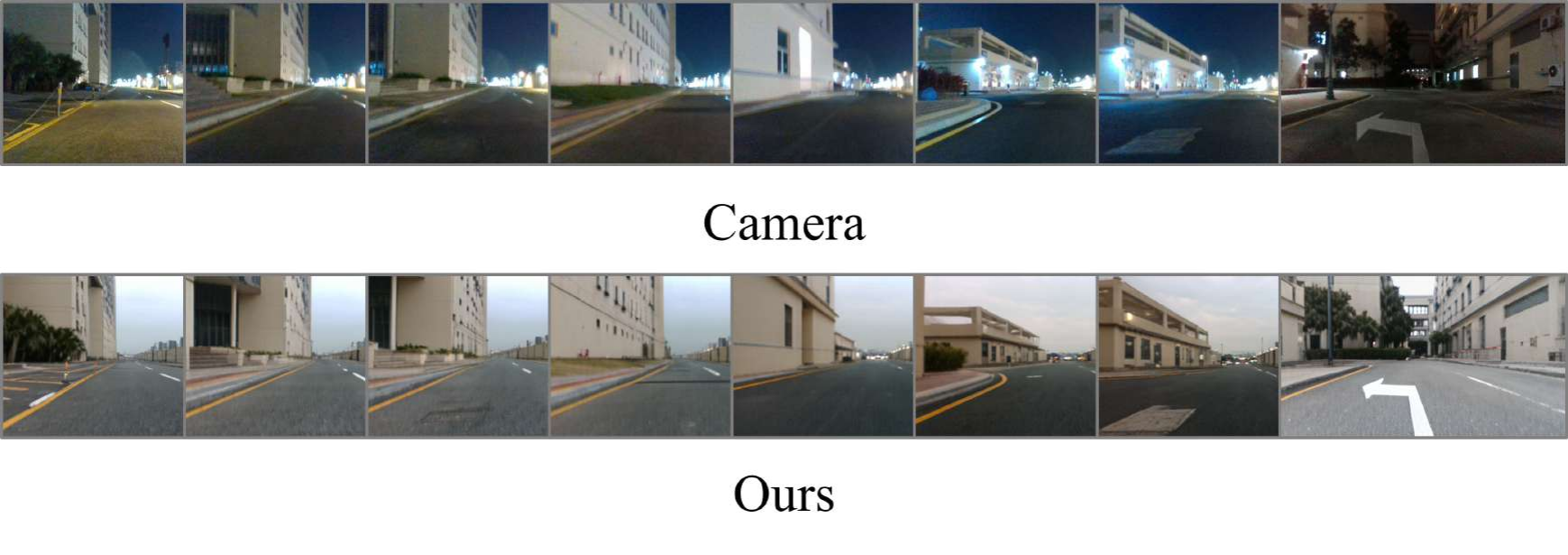}
    \caption{Examples of predicted RGB generated by our method on a low-light sequence.}
    \label{fig:lowlight}
\end{figure}

\begin{figure}
    \centering
    \includegraphics[width=\linewidth]{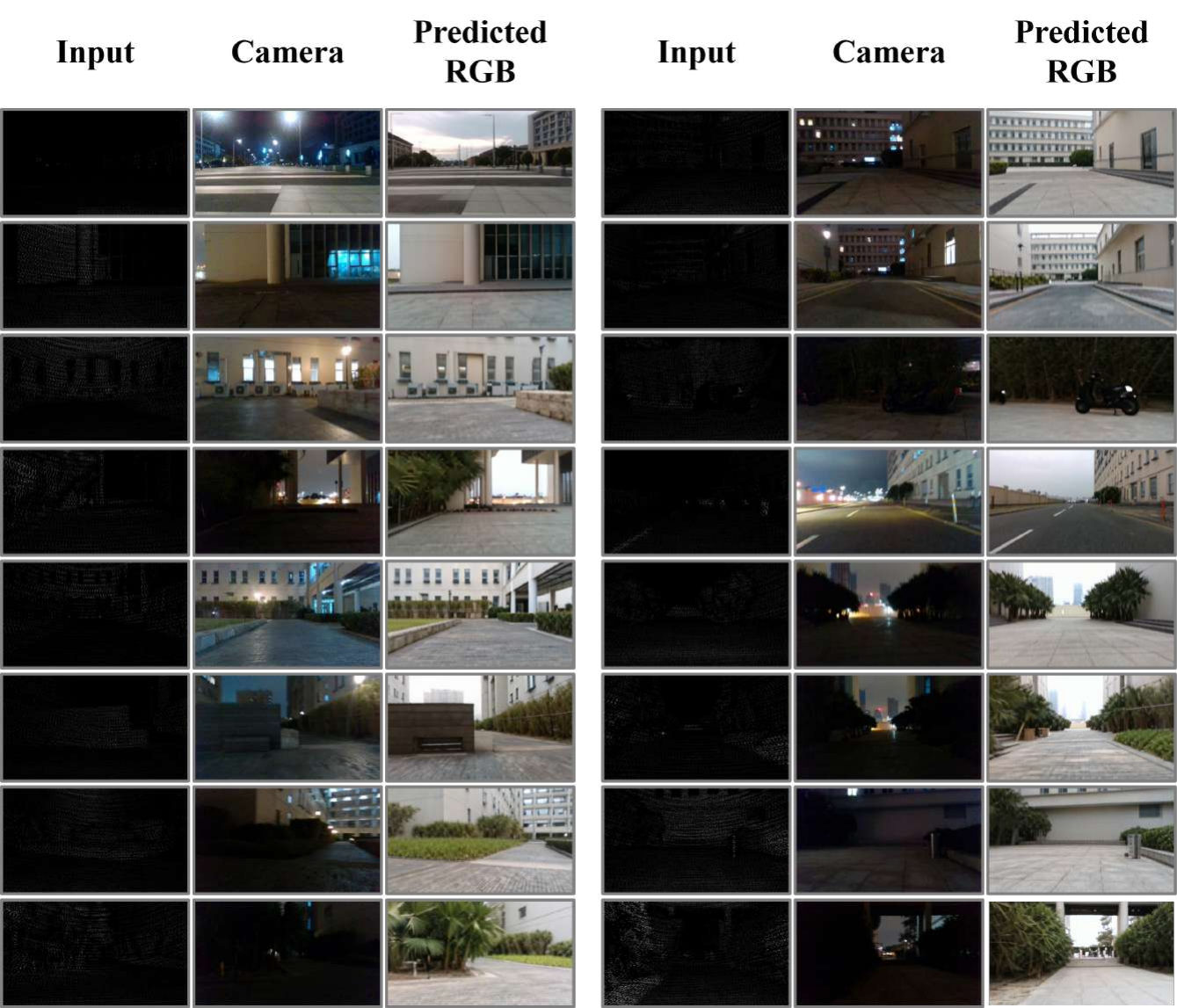}
    \caption{Examples demonstrating that Cyclops produces stable geometry-conditioned RGB renderings across varying illumination conditions, enabling robust all-day visual perception.}
    \label{fig:illumination_invariant}
\end{figure}


\end{document}